\documentclass[journal=jacsat,manuscript=article]{achemso}

\usepackage[version=3]{mhchem} 
\usepackage{hyperref}
\usepackage[inkscapelatex=false]{svg}
\usepackage[utf8]{inputenc} 
\usepackage[T1]{fontenc}    
\usepackage{url}            
\usepackage{booktabs} 
\usepackage{subcaption}
\usepackage{amsfonts}      
\usepackage{nicefrac}       
\usepackage{microtype}      
\usepackage{xcolor}         
\usepackage{graphicx}
\usepackage{amsmath}
\usepackage{amssymb}
\usepackage{geometry}
\usepackage{enumitem}
\usepackage{tcolorbox}
\usepackage{ulem}
\usepackage{tabularx}
\usepackage{listings}
\newcommand{\maxscore}[1]{\underline{#1}}

\definecolor{codegray}{rgb}{0.5,0.5,0.5}
\definecolor{codeblue}{rgb}{0.0,0.2,0.7} 
\definecolor{codegreen}{rgb}{0.2,0.5,0.2} 
\definecolor{codered}{rgb}{0.6,0.0,0.0} 
\definecolor{backcolour}{rgb}{0.98,0.98,0.98} 
\definecolor{framecolour}{rgb}{0.85,0.85,0.85} 

\lstdefinestyle{python-sleek}{
    backgroundcolor=\color{backcolour},   
    commentstyle=\itshape\color{codegreen},
    keywordstyle=\bfseries\color{codeblue},
    numberstyle=\tiny\color{codegray},
    stringstyle=\color{codered},
    basicstyle=\ttfamily\small, 
    breakatwhitespace=true,         
    breaklines=true,                 
    captionpos=b,                    
    keepspaces=true,                 
    numbers=none,                    
    numbersep=5pt,                  
    showspaces=false,                
    showstringspaces=false,
    showtabs=false,                  
    tabsize=2, 
    language=Python,
    frame=single,                    
    rulecolor=\color{framecolour},   
    frameround=tttt,                 
    aboveskip=1em,                   
    belowskip=1em                    
}

\author{Daniel Armstrong}
\author{Zlatko Jončev}
\affiliation[\'Ecole Polytechnique F\'{e}d\'{e}rale de Lausanne (EPFL)]
{\'Ecole Polytechnique F\'{e}d\'{e}rale de Lausanne (EPFL)}

\author{Andres Bran}
\affiliation[\'Ecole Polytechnique F\'{e}d\'{e}rale de Lausanne (EPFL)]
{\'Ecole Polytechnique F\'{e}d\'{e}rale de Lausanne (EPFL)}
\alsoaffiliation[National Centre of Competence in Research (NCCR) Catalysis]
{National Centre of Competence in Research (NCCR) Catalysis}

\author{Philippe Schwaller}
\email{{daniel.armstrong,philippe.schwaller}@epfl.ch}
\affiliation[\'Ecole Polytechnique F\'{e}d\'{e}rale de Lausanne (EPFL)]
{\'Ecole Polytechnique F\'{e}d\'{e}rale de Lausanne (EPFL)}
\title[SynthStrategy]
  {SynthStrategy: Extracting and Formalizing Latent Strategic Insights from LLMs in Organic Chemistry}


\begin{document}

\begin{abstract}
Modern computer-assisted synthesis planning (CASP) systems show promises at generating chemically valid reaction steps but struggle to incorporate strategic considerations such as convergent assembly, protecting group minimization, and optimal ring-forming sequences. We introduce a methodology that leverages Large Language Models to distill synthetic knowledge into code. Our system analyzes synthesis routes and translates strategic principles into Python functions representing diverse strategic and tactical rules, such as strategic functional group interconversions and ring construction strategies. By formalizing this knowledge as verifiable code rather than simple heuristics, we create testable, interpretable representations of synthetic strategy. We release the complete codebase and the USPTO-ST dataset -- synthesis routes annotated with strategic tags. This framework unlocks a novel capability for CASP: natural language-based route retrieval, achieving 75\% Top-3 accuracy on our benchmark. We further validate our library through temporal analysis of historical trends and chemically intuitive route clustering that offers more granular partitioning than common previous methods. This work bridges the tactical-strategic divide in CASP, enabling specification, search, and evaluation of routes by strategic criteria rather than structure alone.
\end{abstract}

\section{Introduction}

Computer-assisted synthesis planning (CASP) has evolved from early rule-based systems to sophisticated machine learning models capable of proposing retrosynthetic disconnections for complex molecules \citep{corey1969computer, Segler2017neural, coley2017prediction, Jin2017predicting,coley2017computer, Schwaller2018found, coley2018, schwaller2020predicting, mo2021evaluating, Sacha2021molecule,irwin2022chemformer,  maziarz2024chimera}. These systems can systematically explore vast chemical spaces to identify pathways connecting target molecules to commercially available starting materials \citep{segler2018planning,grzybowski2018chematica, chen2020retro, genheden2020aizynthfinder, molga2021chemist,genheden_bjerrum_2022, maziarz2023re,tu2025askcos}.

However, a critical gap remains: while these systems typically generate chemically valid steps, they struggle to evaluate routes based on strategic considerations, such as convergent assembly, protecting group minimization, and optimal ring-forming sequences \citep{corey1967general,corey1969computer}. This creates a needle in the haystack problem. CASP can generate thousands of valid routes but cannot identify those with sound strategic design, leading to work on route ranking by neural nets and synthesis cost estimates \citep{mo2021evaluating,badowski2019selection, genheden_bjerrum_2022}. Recent efforts have addressed this challenge with specific synthetic constraints, such as reaction class guidance \citep{toniato2023enhancing}, disconnection prompts \citep{thakkar2023unbiasing}, bond constraints \citep{westerlund2024constrained}, and starting material constraints \citep{yu2024double, armstrong2024tango}. In addition, numerous approaches have attempted to combine a variety of networks to assign a certainty to how likely a synthesis pathway is to work based on single-step reaction scores \citep{segler2018planning, schwaller2020predicting}. Alternative approaches to multi-step route evaluation include statistical methods, which quantifies plausibility through template sequence overlap with known pathways \citep{li2024retro}, and composite scoring schemes that penalize route length while incorporating step confidence and intermediate complexity \citep{kreutter2023tripletransformerloop}. These approaches focus on individual tactical decisions rather than the holistic, high-level reasoning that expert chemists employ when designing synthesis routes. In contrast, to capture more comprehensive strategies, recent work has turned to transformer-based models that autoregressively generate entire synthesis routes, implicitly incorporating multi-step synthetic strategy  \citep{shee2024directmultistep, xuan2025tempre}. Additionally, \citet{roh2025higher} reformulate synthesis planning to align with organic chemistry teaching practice by predicting synthons, enabling the model to focus on higher-level strategic reasoning.

Over the past years, substantial progress has been made in the capabilities of modern Large Language Models (LLMs), numerous studies showing this transfers to a multitude of chemical tasks. Initial demonstrations of chemical knowledge were completed by \citet{jablonka2023gpt} and \citet{llms_for_chem_Guo}, while seminal work by \citet{bran2023chemcrow} and \citet{boiko2023autonomous} revealed the promising performance of GPT-4 in chemical task planning and tool use. Since then, a significant body of work has built up which leverage LLMs for data extraction, retrosynthesis, chemical optimisation and a host of other tasks \citep{jablonka202314, rankovic2023bochemian,rankovic2025gollum,wang2025llm,song2025aot, schilling2025text, narayanan2025training}. Recent work has demonstrated that LLMs possess extensive strategic knowledge, with \citet{bran2025chemical} successfully using LLMs to re-rank CASP outputs based on strategy-specific prompts. However, this approach requires computationally expensive real-time inference for each route. Earlier efforts using tree-LSTMs and similarity metrics \citep{mo2021evaluating, genheden2021clustering, genheden2025simple, genheden2025evaluation} similarly focused on post-hoc evaluation rather than extracting reusable strategic knowledge.
This motivates our central question: if LLMs possess strategic knowledge, can we systematically extract and distill it into a persistent, reusable form that avoids expensive per-route inference?

Inspired by model distillation, where LLMs are used to synthesise data for fine-tuning smaller models on a specific task \citep{gou2021knowledge}, we propose a novel agentic approach: using LLMs' code-generation capabilities \citep{chen2021evaluating, li2023starcoder, austin2021program, google_gemini, anthropic2024claude35sonnet} to translate tacit strategic knowledge into explicit, executable Python functions. Code provides a unique representation: it is interpretable, verifiable through automated testing, composable, and cheap to evaluate at scale. 

Our primary contributions are: first, natural language-driven route retrieval for CASP, achieving 75\% Top-3 accuracy on our benchmark. Second, the USPTO-ST dataset: synthesis routes annotated with strategic tags, enabling improved route retrieval and strategy-aware CASP systems. Third, chemically intuitive route clustering that offers more granular partitioning than common previous methods such as topological metrics. Finally, a curated library of 1,076 strategy functions derived via a methodology combining human and LLM-guided analysis, representing the first large-scale computational distillation of tacit synthetic strategy.

\begin{figure}
   \centering
   \includegraphics[width=0.95\textwidth]{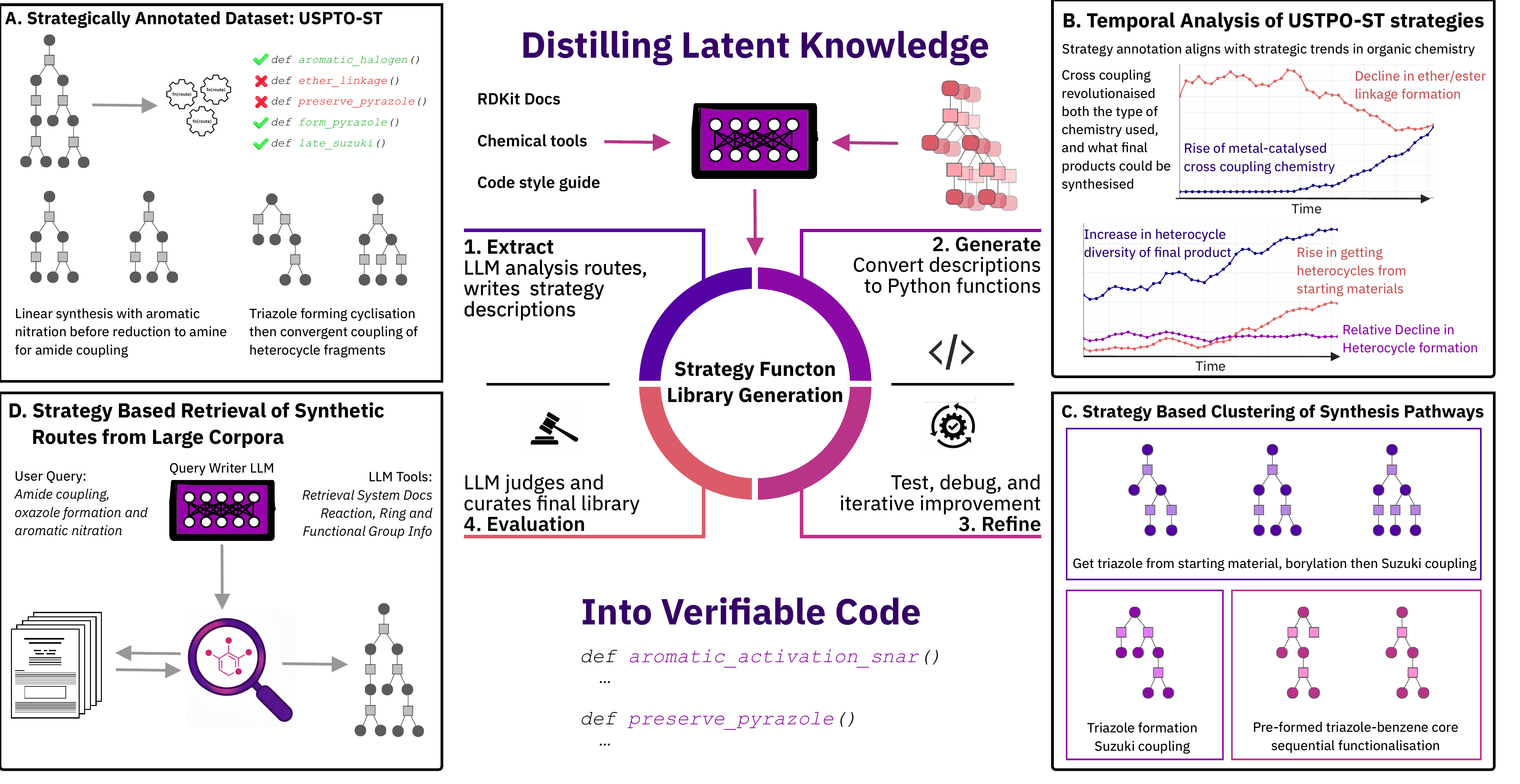}
   \caption{Overview of the SynthStrategy framework for distilling latent chemical knowledge from large language models into a verifiable library of strategy functions. The process involves extracting strategy descriptions from LLM analysis of synthesis routes, generating executable Python functions, and refining them through testing and iteration. Applications include: (A) Creation of the strategically annotated USPTO-ST dataset; (B) Temporal analysis of evolving synthetic strategies in the USPTO dataset; (C) Strategy-based clustering of synthesis pathways for chemically intuitive grouping; (D) Natural language-based retrieval of strategy-matched routes from large corpora.}
   \label{fig:main_fig}
\end{figure}

Our SynthStrategy framework addresses practical challenges (Figure \ref{fig:main_fig}): navigating large solution spaces from CASP tools, communicating strategic preferences, and learning from synthetic precedent. Chemists can query for routes matching strategic criteria (e.g., late-stage functionalization with minimal protecting groups) or cluster large sets of proposed syntheses into chemically meaningful families, overcoming the tendency of purely topological metrics to group strategically distinct routes into single, uninformative clusters. For developers, our dataset provides benchmarks for strategy-aware planning beyond route validity metrics.

Our work addresses the longstanding gap between the local tactical  capabilities (generating plausible reaction steps) of retrosynthesis tools and global strategic reasoning required for practical planning. This gap is enforced as most retrosynthesis models rely on single-step representations like reaction templates, which are fundamentally incapable of capturing the multi-step, relational logic that defines a true synthetic strategy. They cannot encode concepts like convergent assembly, sequential redox manipulations, or protecting group economy. To bridge this tactical-strategic divide, we introduce a new paradigm: distilling implicit strategic knowledge from LLMs into explicit, executable Python functions. Our function library, while not exhaustive, provides a substantial codification of diverse strategic principles, and our USPTO-ST dataset enables rigorous benchmarking of future strategy-aware planning systems. The applications we demonstrate, temporal trend analysis, semantically meaningful clustering, and natural language retrieval, illustrate how explicit strategic representations can enhance both the development and use of CASP tools. This approach enables future systems that integrate strategic heuristics into search algorithms, compose functions for multi-objective constraints, and provide interpretable explanations grounded in explicit strategic principles.



\section{Methods}
\subsection{From single-step representations to executable strategies}

While reaction templates, such as reaction SMARTS, have been foundational for single-step retrosynthesis prediction, they are inherently limited to representing individual, atom-mapped transformations. Consequently, they cannot capture the multi-step, relational logic that defines a synthetic strategy. Our work addresses this gap by representing strategic knowledge as executable functions. Unlike static templates, these functions can analyze the entire synthesis routes, enabling the codification of complex, multi-step sequences. For instance, a function can verify a specific redox manipulation tactic (e.g., an ester reduction followed by a later alcohol oxidation) or a masked amine strategy involving a multi-step azide intermediate. Furthermore, they can evaluate global route properties such as topological convergency, late-stage functionalization, or the overall usage of protecting groups -- concepts beyond the scope of a single-step representation.

A key aspect of our methodology is the use of an LLM to distill which multi-step patterns are strategically meaningful. A brute-force search for all possible reaction sequences would be computationally intractable and yield countless irrelevant correlations. By leveraging the LLM as an expert filter, we focus on codifying the recurrent, coherent strategies employed in practice, distinguishing meaningful plans from incidental reaction noise. This shift from static, single-step representations to dynamic, executable logic allows for a more holistic and chemically nuanced evaluation of synthesis plans.

\subsection{Datasets}
\label{datasets}
We used a cleaned version of the USPTO-STEREO dataset curated by \citet{xuan2025tempre} of chemical reactions originally extracted from the United States Patent and Trademark Office \cite{lowe2012extraction, schwaller2019molecular}. The single step reaction dataset was then converted into multi-step synthesis pathways using a depth first search (DFS).

In addition, we used a common benchmark dataset, PaRoutes for test cases \citep{genheden_bjerrum_2022}. PaRoutes has two formats: n1, which ensures that only one route from any given patent may appear in the test set, and n5, which allows up to five routes from the same patent in the test set \citep{genheden_bjerrum_2022}. Details on the number of reactions and routes in each set are given in the Appendix.

We created a strategy extraction benchmark by taking 2,500, 500 and 1000 routes from the USPTO training and validation sets and the PaRoutes-n1 test set, respectively. We used PaRoutes n1 to increase the strategic diversity of routes analyzed. For the route clustering experiment we took the 5,000 molecules from ChEMBL used by \citet{genheden2021clustering} and ran AiZynthFinder to generate routes.

For our strategy-based route retrieval task, we created a human-curated benchmark from routes in the PaRoutes-n5 testset. We select a pool of candidate routes based on the following criteria: 
\begin{itemize}
    \item Complexity: routes which have an unusually high number of strategic functions,
    \item Categorical Checks: routes which contain unique or rare categorical checks (named reactions, ring systems and functional groups,
    \item Strategic Intricacy: routes which contain uncommon combinations of functions frequent in the entire corpus.
\end{itemize}

This led to a benchmark containing 55 test cases.

\paragraph{Automated Generation of a Chemical Strategy Library}
We used LLMs (Claude-3.7-Sonnet) to generate a library of Python functions, each representing a specific strategic or tactical rule. We constructed this library via a multi-stage knowledge distillation process using LLMs. A LLM is a large neural network trained to understand complex information, perform reasoning tasks, and generate new content, such as computer code \citep{openai_gpt4, anthropic2024claude35sonnet}. In our process, an LLM analysed a synthesis route, abstracted the underlying chemical strategy, and translated it into a corresponding, verifiable Python function.

We prompted our LLM with synthesis pathways and asked it to: (1) articulate the underlying strategy in natural language, and (2) translate this into an executable Python function. For this stage, we used Claude-3.7-Sonnet, which we found to be the most capable model at the time. Each route came from a unique patent leading to a more strategically rich dataset. The functions were designed to return True if a given synthesis route satisfied the specific strategic criterion. To aid in this task, the LLM was equipped with a comprehensive suite of cheminformatics tools, including RDKit and custom functions for analyzing molecular structures, reactions, and synthesis trees. These custom functions included checkers for named reactions, functional groups, and ring systems, leveraging reaction templates and SMARTS patterns from the Rxn-Insight toolkit \citep{rxn-insight}.

This initial set of generated functions underwent an automated refinement pipeline. Functions were tested for correctness against the synthesis route for which they were generated. The functions that failed were iteratively refactored by the LLM based on error messages and standard output. This process again used Claude-3.7-Sonnet. Subsequently, the functions were further refined and evaluated for chemical relevance and quality by newly released more advanced LLMs (Gemini-2.5-Flash and Gemini-2.5-Pro, as they had been shown to have better chemical reaction and route understanding in the benchmark by \citet{bran2025chemical}). We used strict constraints that preserved the functions' core logic. This iterative cycle of generation, testing, and constrained refactoring yielded our final library of 1,076 functions. Further details on this process are given in the Appendix \ref{app:xtended_methods}. \ref{app:refactoring_frameworks}.

\paragraph{Strategy Based Clustering of CASP tool outputs}
To understand the fundamental patterns of synthesis strategies represented in our function library, we performed a clustering analysis. We applied our library of functions to 5,000 synthesis routes planned for molecules from the ChEMBL dataset. Each route was converted into a binary strategy fingerprint, where each position indicates whether a specific strategy function was satisfied. Using the K-Means clustering algorithm, we grouped routes with similar fingerprints. This allowed us to identify distinct clusters, or strategies each defined by a coherent set of co-occurring functions, revealing the common combinations of tactics used in synthetic planning.

\paragraph{Natural Language-Based Strategy Retrieval}
Building on our function library, we created a retrieval framework that enables chemists to search for synthesis routes using complex, descriptive queries. A user can describe a desired strategy in natural language (e.g., "a convergent synthesis that forms a key C-C bond using a named reaction late-stage"). A Query Rewriter LLM (Gemini-2.5-Flash) translates this input into a structured query composed of the underlying strategic functions. The system then efficiently searches using joint semantic and categorical matching to find routes that match this combination of functions. Semantic matching estimates text similarity between the user's query and the function description, while Categorical Checks test for the presence of specific named reactions, functional groups or rings. The retrieved routes are ranked based on how completely and accurately they satisfy the user's original query, providing a powerful tool for navigating vast spaces of synthetic possibilities.

\section{Results}
\subsection{Unlocking Natural Language-Based Synthesis Route Retrieval}
The most transformative application of our function library is the ability to navigate large synthesis corpora using high-level strategic concepts rather than sub-structure matching. We developed a retrieval framework that translates natural language queries (e.g., convergent synthesis with late-stage Suzuki coupling) into structured searches over our function library. This tackles the challenge where chemists must find strategically relevant routes among thousands of valid but generic possibilities. 

\begin{figure}
   \centering
   \includegraphics[width=0.95\textwidth]{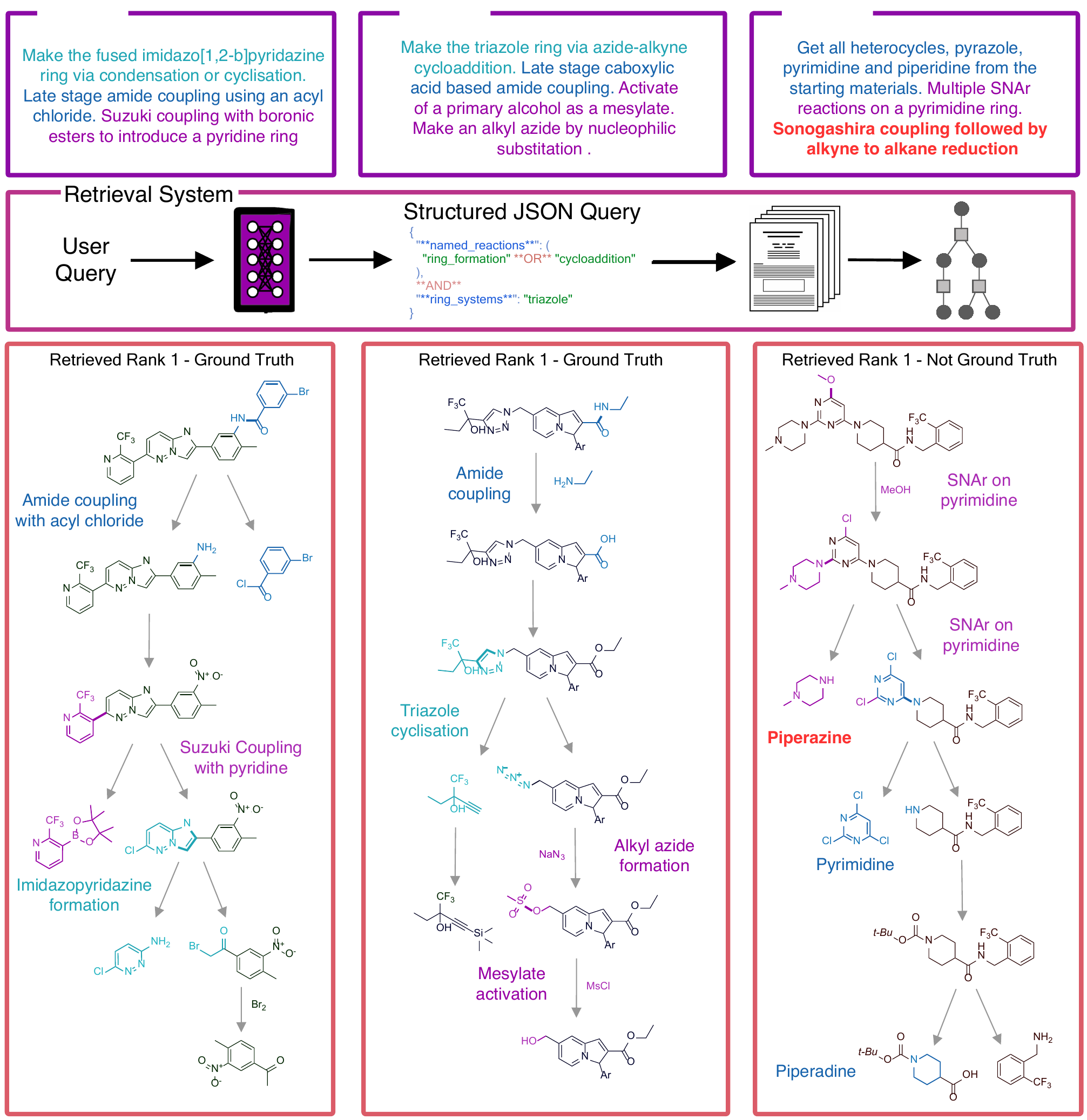}
   \caption{Here we show three examples from our retrieval benchmark. In Queries A and B, the ground truth routes are returned as the Rank 1 result. For Query C, the retrieval system fails to correctly find the desired route at Rank 1, failing to match the request for a route which preserves a pyrazole from the starting materials ( instead finding a match for 2 out of 3 desired heterocycles ) and uses a Sonogashira coupling to introduce an alkyl linker via alkyne reduction. However, the system does find the ground truth route at Rank 3. }
   \label{fig:retrieval_case_study}
\end{figure}

Our system operates in three phases. First, an LLM transforms the user's input into a structured query containing sub-queries with both semantic descriptions and categorical checks (e.g., specific named reactions or ring systems). Second, the system queries the function library. This step can utilize semantic similarity to filter candidate functions, strict categorical checks for hard filtering, or a hybrid of both. Finally, synthesis routes are retrieved, filtered, and ranked based on their satisfaction of these executable functions.

We evaluated this system on a human-curated benchmark of 55 complex routes from the PaRoutes-n5 test set \cite{genheden_bjerrum_2022}. Comparing embedding architectures (Table~\ref{tab:model_accuracy_comparison}), we found that smaller, open-source models like MiniLM-L6 (22.7M parameters) perform competitively against significantly larger, closed-source models like OpenAI's ada-002 (50.9\% vs 52.7\% Top-1 in the default setting). This indicates that retrieval performance is driven primarily by the quality of the distilled strategy library rather than the raw power of the embedding model.

To disentangle the contributions of the search components, we performed an ablation study using MiniLM-L6 (Table~\ref{tab:model_accuracy_comparison}, bottom). The results highlight the critical importance of categorical precision over semantic approximation for this task. Relying solely on semantic matching is ineffective (3.6\% Top-1). Furthermore, the "Default" pipeline -- which uses semantic similarity to filter functions \textit{before} applying categorical checks -- underperforms compared to strict categorical approaches (50.9\% vs 60.0\% Top-1). 

We hypothesize that on this concise benchmark, initial semantic filtering acts as an aggressive pruning step, inadvertently discarding functions that are strategically correct but semantically distant from the query. Consequently, the "Categorical Only" approach achieves the highest Top-1 accuracy (60.0\%) by avoiding this lossy pre-filtering. However, semantic information remains valuable for ranking; the hybrid "Cat + Sem Rerank" approach (categorical filtering followed by semantic sorting) recovers the highest Top-3 accuracy (74.6\%), effectively breaking ties among categorically valid routes. While strict categorical matching dominates this benchmark, semantic filtering may still be necessary for computational efficiency when scaling to massive, noisy datasets.

\begin{table}
\centering
\caption{Top-K accuracy on the retrieval benchmark by embedding model. Joint highest scores are underlined. Ablations: (Default) semantic filtering followed by categorical checks; (Cat Only) strict categorical matching only; (Cat + Sem Rerank) categorical matching followed by semantic re-ranking; (Sem) semantic matching only.}
\label{tab:model_accuracy_comparison}
\begin{tabularx}{\columnwidth}{l>{\raggedleft\arraybackslash}p{1.3cm}*{4}{>{\centering\arraybackslash}X}}
\toprule
\textbf{Model} &\textbf{Param.} \newline \textbf{Count} & \textbf{top-1 (\%)} & \textbf{top-3 (\%)} & \textbf{top-5 (\%)} & \textbf{top-10 (\%)} \\
\midrule
RoBERTa Large & 355M & 20.0 & 36.4 & 45.5 & 50.9 \\
SciBERT & 110M & 23.6 & 29.1 & 38.2 & 58.2 \\
Qwen2 4B & 4000M & 30.9 & 50.9 & 50.9 & 63.6 \\
Qwen2 1.5B & 1500M & 30.9 & 40.0 & 54.5 & 60.0 \\
Qwen2 0.6B & 600M & 38.2 & 58.2 & 61.8 & 65.5 \\
OpenAI 3 Large & -- & 49.1 & 65.5 & 67.3 & 72.7 \\
OpenAI ada-002 & -- & 52.7 & 63.6 & 67.3 & 70.9 \\

\midrule
\multicolumn{6}{l}{\textbf{MiniLM-L6 Ablations}} \\
MiniLM-L6 (Default) & 22.7M & 50.91 & 60.00 & 67.27 & 70.91 \\
MiniLM-L6 (Cat Only) & 22.7M & \maxscore{60.00} & 72.73 & 76.36 & \maxscore{80.00} \\
MiniLM-L6 (Cat + Sem Rerank) & 22.7M & 58.18 & \maxscore{74.55} & \maxscore{78.18} & \maxscore{80.00} \\
MiniLM-L6 (Sem) & 22.7M & 3.64 & 10.91 & 12.73 & 14.55 \\
\bottomrule
\end{tabularx}
\end{table}

\paragraph{Validation with historical trends}
To validate our methodology, we tested whether our functions capture known historical trends in synthetic chemistry. We applied our complete function library to the entire USPTO dataset, which spans several decades of patented chemical syntheses, and aggregated the prevalence of key strategic classes over time. To do this, we manually identified a relevant function for specific strategies and, for each year, calculated the fraction of synthetic routes for which each function returned True.

\begin{figure}
\centering

   \includegraphics[width=0.95\textwidth]{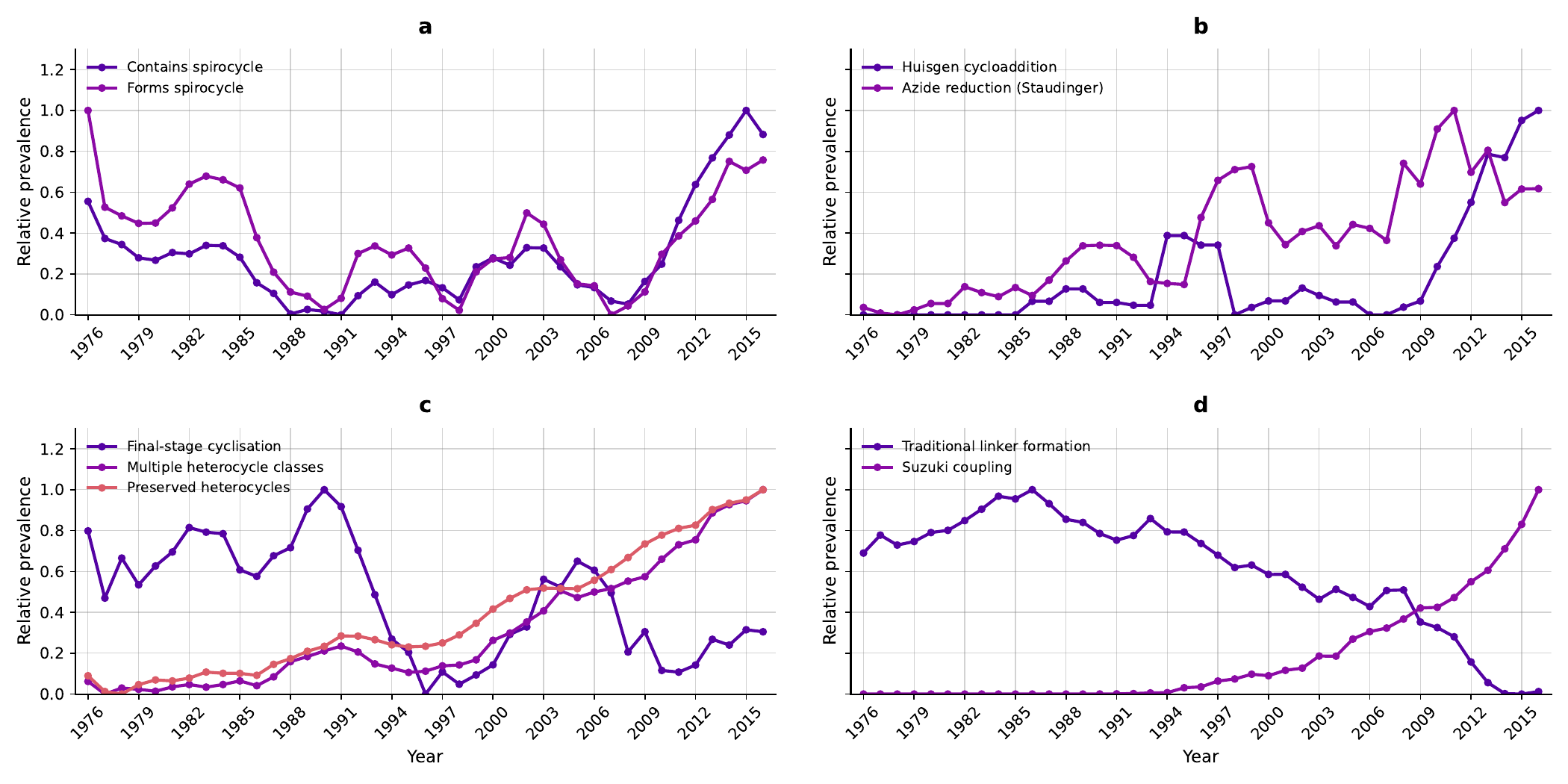}
\caption{
    Temporal analysis of synthetic strategies in the USPTO dataset. 
    \textbf{(a)} The presence versus formation rates of spirocyclic compounds in USPTO routes has seen a marked increase over the past 5 years. 
    \textbf{(b)} In orange we see the Azide to amine Staudinger reduction while in magenta, the rise in use of 1,3 cycloadditions to form triazoles is observed post 2007.
    \textbf{(c)} The percentage of routes in which a function representing heterocycle preservation from the starting material returns True has steadily increased, while strategies involving de novo ring construction via cyclization have stagnated. 
    \textbf{(d)} This trend is mirrored by the dramatic rise of Suzuki Coupling, which facilitates the connection of pre-built molecular fragments, coinciding with the relative decline of traditional linker reactions.
}
\label{fig:temporal_trends}
\end{figure}

The results, shown in Figure \ref{fig:temporal_trends}, reveal clear and well-documented strategic shifts. First, we observe specific trends in structural complexity and reactivity. Figure \ref{fig:temporal_trends}a illustrates a marked increase in the presence of spirocyclic compounds over the last five years, consistent with the recent medicinal chemistry focus on increasing three-dimensional character in drug candidates. Similarly, Figure \ref{fig:temporal_trends}b captures the evolution of azide chemistry: while the classical Staudinger reduction to amines remains constant (orange), there is a distinct emergence of 1,3-cycloadditions post-2007 (magenta), reflecting the widespread adoption of "click chemistry" for rapid triazole synthesis.

Broader shifts in molecular assembly are also evident. Figure \ref{fig:temporal_trends}c shows a steady rise in routes that preserve pre-existing heterocycles from starting materials, while strategies involving de novo ring construction via cyclisation have stagnated. This indicates a strategic shift towards purchasing heterocycle-containing building blocks and joining them together. We hypothesize that this trend is largely driven by the ascendancy of modern cross-coupling reactions, which excel at stitching complex fragments together. Figure \ref{fig:temporal_trends}d strongly corroborates this hypothesis, charting the dramatic rise of the Suzuki Coupling, a Nobel Prize-winning reaction that has become a dominant tool for C–C bond formation. Its growth directly coincides with the relative decline of more traditional fragment-linking reactions such as etherification and esterification.

\subsection{Chemically Intuitive Route Clustering}
Computer-Aided Synthesis Planning (CASP) tools often generate numerous potential routes to a target molecule, creating a significant challenge in analysis and selection. A common approach to differentiate these routes is clustering based on topological similarity, such as the tree-edit distance (TED) between synthesis graphs \citep{genheden2021clustering}. However, such methods are agnostic to the underlying chemical logic, potentially grouping strategically dissimilar routes. We propose that by using our library of explicit strategy functions as route featurizers, we can achieve a more granular and chemically meaningful partitioning of the solution space. We provide further details on the clustering approach in the Appendix.

\begin{figure}
   \centering
   \includegraphics[width=0.95\textwidth]{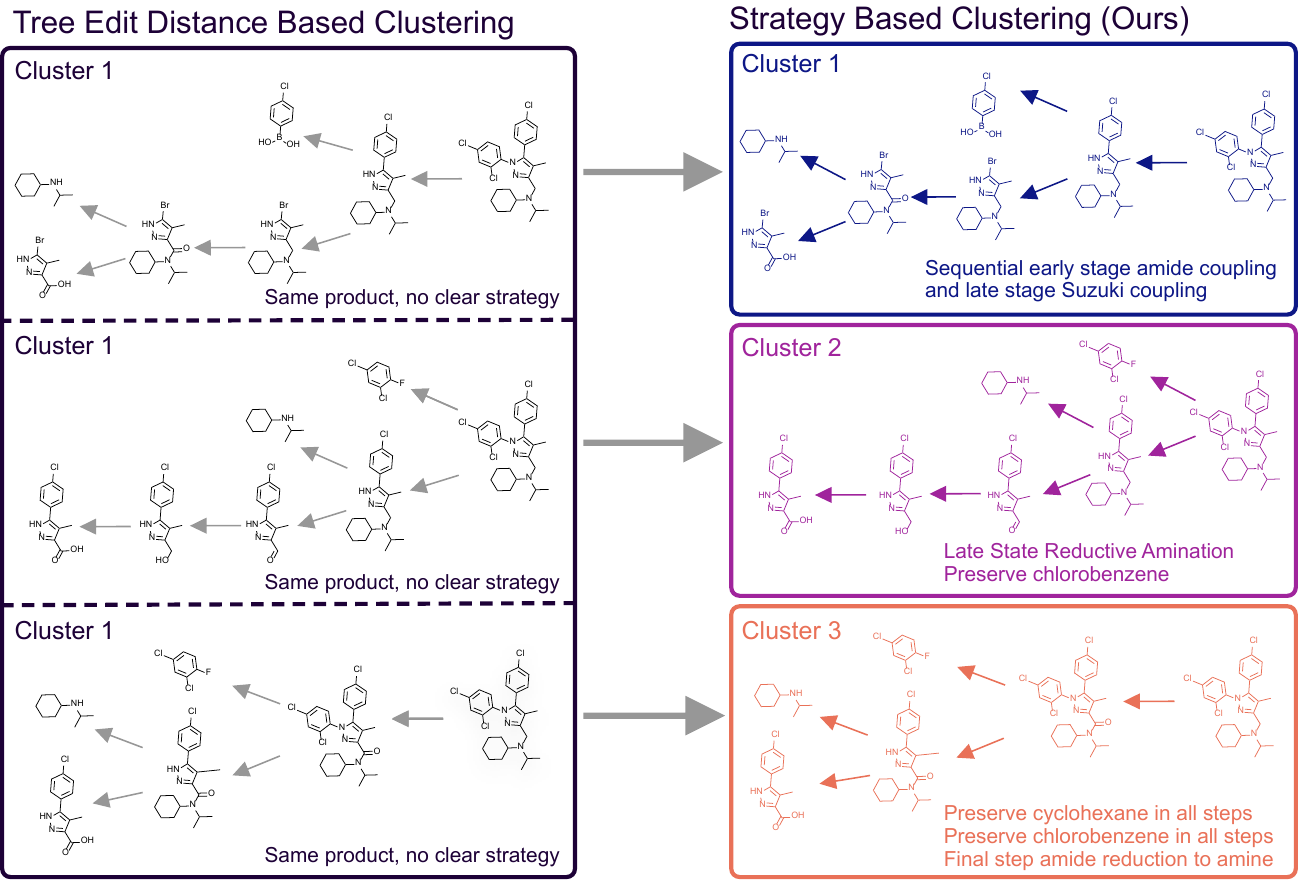}
   \caption{Here we show cluster representatives of clustering synthesis pathways to an ORL1 Antagonist. Strategy cluster representatives are numbered and grouped within boxes. TED clusters are denoted by colour. Strategy Clustering used KMeans and returned K=5 and TED Clustering used Agglomerative clustering and returned K=3 for an optimal number of clusters. We note that out of seven total routes were generated by AiZynthFinder, but for clarity not all are shown here.}
   \label{fig:case_study}
\end{figure}

We hypothesize that strategic functions will cluster routes by synthetic logic rather than just structural similarity, providing chemically meaningful groupings that topological metrics miss. The utility of this approach is demonstrated in a case study on routes generated for an ORL1 antagonist (Figure~\ref{fig:case_study}). As shown, TED-based clustering groups three distinct synthetic plans into a single, undifferentiated cluster. While structurally related, these routes employ fundamentally different strategies that are not captured by the topological metric.
In stark contrast, our strategy-based method partitions these same three routes into separate, chemically interpretable clusters, each defined by a clear synthetic logic:
\begin{itemize}
\item \textbf{Cluster 1:} A convergent route featuring an early-stage amide coupling followed by a late-stage Suzuki coupling.
\item \textbf{Cluster 2:} A strategy characterized by a late-stage reductive amination, while preserving a chlorobenzene motif throughout.
\item \textbf{Cluster 3:} A linear approach where both cyclohexane and chlorobenzene motifs are preserved, culminating in a final-step amide reduction.
\end{itemize}
This example demonstrates that our method successfully distinguishes between routes that, while topologically similar enough to be grouped by TED, employ different high-level plans. Our approach identifies crucial strategic differences, such as the sequence of key bond formations and scaffold preservation tactics, which are overlooked by purely structural metrics.

We provide additional quantitative analysis of cluster size and the variation in cluster sizes in Appendix \ref{app:strategy_clustering} and Figure \ref{fig:clustering_comparison}. In general, strategy clusters produce a larger number of clusters with lower variance in cluster size, mitigating the tendency of previous methods such as TED to produce a very small number of clusters for a large number of routes.

\section{Discussion}

The methodology presented here addresses the disconnect between the tactical capabilities of modern CASP systems and the strategic needs of practicing chemists. While existing models excel at proposing valid chemical steps, they often fail to organize these steps into a coherent, purposeful plan. By leveraging LLMs to distill implicit synthetic knowledge into explicit, executable Python functions, we establish a new paradigm for representing chemical strategy -- one that is verifiable, scalable, and interpretable.

\paragraph{Code as a bridge between implicit and explicit knowledge}
The core innovation of this work is the formalization of strategy as \textit{code}. Unlike vector embeddings or text descriptions, code serves as a unique bridge: it captures the nuance of the LLM's understanding but freezes it into a deterministic, testable form. As demonstrated by the representative functions in Figure~\ref{fig:code_examples}, this allows for the encoding of complex, multi-step logic -- such as temporal constraints in redox manipulations or "masked" functional group strategies -- that single-step representations like reaction SMARTS are fundamentally incapable of capturing.

\paragraph{Efficiency and scalability via distillation}
A critical advantage of our approach is recording the knowledge of LLMs in a concrete format. Recent efforts have demonstrated that LLMs can act as effective critics for synthesis plans \citep{bran2025chemical}, but this requires expensive, high-latency inference for every route evaluated. Our framework acts as a knowledge distillation pipeline: we pay the high computational cost of the LLM only once during the \textit{generation} of the function library. Once formalized, these Python functions can be applied to millions of routes at negligible cost. Particularly important is the ability to perform reliable pairwise comparisons, which is difficult and expensive with LLMs. This allows for the screening of synthesis pathways that would be economically infeasible with direct LLM prompting.

\paragraph{Interpretability}
The adoption of CASP tools has historically been hindered by the "black box" nature of neural scoring functions. When a model ranks one route over another based on a probability score, the chemist has no insight into the reasoning. In contrast, our framework offers "glass box" interpretability. When our retrieval system identifies a route, or our clustering algorithm groups pathways, the rationale is grounded in explicit logic (e.g., "Function 402: Late-Stage Suzuki" returned True). This transparency is essential for building trust with expert users, who can inspect, verify, and even modify the underlying code to align with their specific constraints.

Our framework is also highly extensible. Functions can be easily added post hoc, either through manual programming, prompting an LLM to provide code for a specified strategy, or by simply running the Extract, Generate, Refine and Evaluate cycle on a synthesis pathway which contains the desired strategy. 

\paragraph{The USPTO-ST Dataset}
Beyond the immediate applications of retrieval and clustering, the release of the USPTO-ST dataset -- synthesis routes annotated with our strategic tags -- provides a foundation for future research. Currently, retrosynthesis models are trained primarily on valid reactions, not valid \textit{strategies}. By providing a dataset where routes are labeled with high-level concepts (e.g., "Convergent," "Protecting-Group-Free"), we enable the training of future supervised models that can predict strategic attributes directly, potentially allowing for the conditioning of generative models on specific strategic goals.


\begin{figure}[ht]
    \centering
    \begin{minipage}[t]{0.46\textwidth}
        \textbf{a.}
        \begin{lstlisting}[style=python-sleek, basicstyle=\ttfamily\tiny]
def main(route: Dict) -> bool:
    """
    Detects the masked amine strategy via an azide intermediate.
    Forward: alcohol -> sulfonate -> azide -> amine -> amide
    Retro: amide -> amine -> azide -> sulfonate -> alcohol
    """
    # Flags to track if each step in the sequence is found
    found = {
        'amide_coupling': False, 'azide_reduction': False,
        'azide_formation': False, 'alcohol_activation': False
    }

    def dfs_traverse(node):
        if node["type"] == "reaction":
            rsmi = node["metadata"]["mapped_reaction_smiles"]
            # Check for each step of the sequence
            if checker.check_reaction("Amide Formation", rsmi):
                found['amide_coupling'] = True
            if checker.check_reaction("Azide to amine reduction (Staudinger)", rsmi):
                found['azide_reduction'] = True
            if checker.check_reaction("Azide Formation from Halogen/Alcohol", rsmi):
                found['azide_formation'] = True
            if checker.check_reaction("Alcohol Activation (e.g., Sulfonate)", rsmi):
                found['alcohol_activation'] = True

        for child in node.get("children", []):
            dfs_traverse(child)

    dfs_traverse(route)

    # Strategy is confirmed only if all four steps are present
    return all(found.values())
        \end{lstlisting}
    \end{minipage}
    \hfill 
    \begin{minipage}[t]{0.46\textwidth}
        \textbf{b.}
        \begin{lstlisting}[style=python-sleek, basicstyle=\ttfamily\tiny]
def main(route: Dict) -> bool:
    """
    Detects a redox sequence: an ester is reduced to an alcohol,
    which is later oxidized to an aldehyde or ketone.
    """
    found_reduction = False
    found_oxidation = False
    reduction_depth = -1
    oxidation_depth = -1

    def dfs_traverse(node, depth=0):
        nonlocal found_reduction, found_oxidation, reduction_depth, oxidation_depth
        if node["type"] == "reaction":
            rsmi = node["metadata"]["rsmi"]

            if checker.check_reaction("Reduction of ester to primary alcohol", rsmi):
                found_reduction = True
                reduction_depth = depth

            if checker.check_reaction("Oxidation of Alcohols to Aldehydes/Ketones", rsmi):
                found_oxidation = True
                oxidation_depth = depth

        for child in node.get("children", []):
            # Depth increases for chemical nodes, not reaction nodes
            new_depth = depth + 1 if node["type"] != "reaction" else depth
            dfs_traverse(child, new_depth)

    dfs_traverse(route)

    # Strategy requires co-occurrence AND correct ordering.
    return found_reduction and found_oxidation and (oxidation_depth < reduction_depth)  



    
        \end{lstlisting}
    \end{minipage}

    \caption{Two representative strategy functions rated `Perfect' by our LLM evaluator. (a) This function detects a four-step sequence for masking an amine as an azide. (b) This function identifies a redox manipulation strategy, ensuring both the presence and correct temporal ordering of an ester reduction and an alcohol oxidation.}
    \label{fig:code_examples}
\end{figure}

\paragraph{Limitations and Future Directions}
Several limitations warrant consideration. First, the reliance on LLM judgments introduces potential biases towards specific chemical strategies from model training data, although our iterative testing pipeline using an alternate language model may mitigate this. Second, the functions are derived primarily from patent data, potentially over-representing industrial-scale strategies at the expense of academic or exploratory approaches. Third, hallucinations in SMARTS pattern generation required strict constraints to the human-curated lists contained in Rxn-Insight \citep{rxn-insight}. The advent of models with stronger SMILES and SMARTS generation ability may allow this constraint to be weakened in the future. We note that newer models (e.g., Gemini-2.5-Pro) show improved chemical reasoning, suggesting that the fidelity of strategy distillation will improve alongside model capabilities. 


\section{Conclusion}

In summary, we have introduced SynthStrategy, a framework that bridges the tactical-strategic divide in computer-assisted synthesis planning by translating implicit chemical knowledge into a library of explicit, executable functions. Our methodology, which uses LLMs to distill strategic route reasoning from patent syntheses into verifiable Python functions, establishes a new, structured paradigm for capturing, testing, and applying strategic knowledge. The resulting strategy library and the USPTO-ST dataset enable quantitative analysis of chemical trends, provide high-granularity strategy-aware clustering, and facilitate the retrieval of syntheses via natural language queries. By formalizing chemical intuition into computationally tractable code, this work provides a direct path toward CASP systems that can be steered, constrained, and understood at a higher level of strategic abstraction, moving the field closer to the reasoning of an expert chemist.

\section{Code and Data Availability}

We provide access to the code base on \href{https://github.com/schwallergroup/synth_strategy.git}{Github} and release the data on \href{https://figshare.com/s/acd0b405ee671d9327a9}{Figshare}. 

\section{Acknowledgments and Disclosure of Funding}
This research was funded by the Swiss National Science Foundation (SNSF) under grant nunber 214915. In addition we thank the National Centre of Competence in Research (NCCR) Catalysis for support under grant number 180544.

\bibliography{bib}

\providecommand{\latin}[1]{#1}
\makeatletter
\providecommand{\doi}
  {\begingroup\let\do\@makeother\dospecials
  \catcode`\{=1 \catcode`\}=2 \doi@aux}
\providecommand{\doi@aux}[1]{\endgroup\texttt{#1}}
\makeatother
\providecommand*\mcitethebibliography{\thebibliography}
\csname @ifundefined\endcsname{endmcitethebibliography}  {\let\endmcitethebibliography\endthebibliography}{}
\begin{mcitethebibliography}{58}
\providecommand*\natexlab[1]{#1}
\providecommand*\mciteSetBstSublistMode[1]{}
\providecommand*\mciteSetBstMaxWidthForm[2]{}
\providecommand*\mciteBstWouldAddEndPuncttrue
  {\def\EndOfBibitem{\unskip.}}
\providecommand*\mciteBstWouldAddEndPunctfalse
  {\let\EndOfBibitem\relax}
\providecommand*\mciteSetBstMidEndSepPunct[3]{}
\providecommand*\mciteSetBstSublistLabelBeginEnd[3]{}
\providecommand*\EndOfBibitem{}
\mciteSetBstSublistMode{f}
\mciteSetBstMaxWidthForm{subitem}{(\alph{mcitesubitemcount})}
\mciteSetBstSublistLabelBeginEnd
  {\mcitemaxwidthsubitemform\space}
  {\relax}
  {\relax}

\bibitem[Corey and Wipke(1969)Corey, and Wipke]{corey1969computer}
Corey,~E.~J.; Wipke,~W.~T. Computer-Assisted Design of Complex Organic Syntheses: Pathways for molecular synthesis can be devised with a computer and equipment for graphical communication. \emph{Science} \textbf{1969}, \emph{166}, 178--192\relax
\mciteBstWouldAddEndPuncttrue
\mciteSetBstMidEndSepPunct{\mcitedefaultmidpunct}
{\mcitedefaultendpunct}{\mcitedefaultseppunct}\relax
\EndOfBibitem
\bibitem[Segler and Waller(2017)Segler, and Waller]{Segler2017neural}
Segler,~M.~H.; Waller,~M.~P. Neural-symbolic machine learning for retrosynthesis and reaction prediction. \emph{Chem. Eur. J.} \textbf{2017}, \emph{23}, 5966--5971\relax
\mciteBstWouldAddEndPuncttrue
\mciteSetBstMidEndSepPunct{\mcitedefaultmidpunct}
{\mcitedefaultendpunct}{\mcitedefaultseppunct}\relax
\EndOfBibitem
\bibitem[Coley \latin{et~al.}(2017)Coley, Barzilay, Jaakkola, Green, and Jensen]{coley2017prediction}
Coley,~C.~W.; Barzilay,~R.; Jaakkola,~T.~S.; Green,~W.~H.; Jensen,~K.~F. Prediction of organic reaction outcomes using machine learning. \emph{ACS Cent. Sci.} \textbf{2017}, \emph{3}, 434--443\relax
\mciteBstWouldAddEndPuncttrue
\mciteSetBstMidEndSepPunct{\mcitedefaultmidpunct}
{\mcitedefaultendpunct}{\mcitedefaultseppunct}\relax
\EndOfBibitem
\bibitem[Jin \latin{et~al.}(2017)Jin, Coley, Barzilay, and Jaakkola]{Jin2017predicting}
Jin,~W.; Coley,~C.; Barzilay,~R.; Jaakkola,~T. Predicting Organic Reaction Outcomes with Weisfeiler-Lehman Network. Advances in Neural Information Processing Systems 30. 2017; pp 2607--2616\relax
\mciteBstWouldAddEndPuncttrue
\mciteSetBstMidEndSepPunct{\mcitedefaultmidpunct}
{\mcitedefaultendpunct}{\mcitedefaultseppunct}\relax
\EndOfBibitem
\bibitem[Coley \latin{et~al.}(2017)Coley, Rogers, Green, and Jensen]{coley2017computer}
Coley,~C.~W.; Rogers,~L.; Green,~W.~H.; Jensen,~K.~F. Computer-assisted retrosynthesis based on molecular similarity. \emph{ACS Cent. Sci.} \textbf{2017}, \emph{3}, 1237--1245\relax
\mciteBstWouldAddEndPuncttrue
\mciteSetBstMidEndSepPunct{\mcitedefaultmidpunct}
{\mcitedefaultendpunct}{\mcitedefaultseppunct}\relax
\EndOfBibitem
\bibitem[Schwaller \latin{et~al.}(2018)Schwaller, Gaudin, Lanyi, Bekas, and Laino]{Schwaller2018found}
Schwaller,~P.; Gaudin,~T.; Lanyi,~D.; Bekas,~C.; Laino,~T. ``{Found in Translation}'': predicting outcomes of complex organic chemistry reactions using neural sequence-to-sequence models. \emph{Chem. Sci.} \textbf{2018}, \emph{9}, 6091--6098\relax
\mciteBstWouldAddEndPuncttrue
\mciteSetBstMidEndSepPunct{\mcitedefaultmidpunct}
{\mcitedefaultendpunct}{\mcitedefaultseppunct}\relax
\EndOfBibitem
\bibitem[Coley \latin{et~al.}(2018)Coley, Green, and Jensen]{coley2018}
Coley,~C.~W.; Green,~W.~H.; Jensen,~K.~F. Machine Learning in Computer-Aided Synthesis Planning. \emph{Accounts of Chemical Research} \textbf{2018}, \emph{51}, 1281--1289\relax
\mciteBstWouldAddEndPuncttrue
\mciteSetBstMidEndSepPunct{\mcitedefaultmidpunct}
{\mcitedefaultendpunct}{\mcitedefaultseppunct}\relax
\EndOfBibitem
\bibitem[Schwaller \latin{et~al.}(2020)Schwaller, Petraglia, Zullo, Nair, Haeuselmann, Pisoni, Bekas, Iuliano, and Laino]{schwaller2020predicting}
Schwaller,~P.; Petraglia,~R.; Zullo,~V.; Nair,~V.~H.; Haeuselmann,~R.~A.; Pisoni,~R.; Bekas,~C.; Iuliano,~A.; Laino,~T. Predicting retrosynthetic pathways using transformer-based models and a hyper-graph exploration strategy. \emph{Chem. Sci.} \textbf{2020}, \emph{11}, 3316--3325\relax
\mciteBstWouldAddEndPuncttrue
\mciteSetBstMidEndSepPunct{\mcitedefaultmidpunct}
{\mcitedefaultendpunct}{\mcitedefaultseppunct}\relax
\EndOfBibitem
\bibitem[Mo \latin{et~al.}(2021)Mo, Guan, Verma, Guo, Fortunato, Lu, Coley, and Jensen]{mo2021evaluating}
Mo,~Y.; Guan,~Y.; Verma,~P.; Guo,~J.; Fortunato,~M.~E.; Lu,~Z.; Coley,~C.~W.; Jensen,~K.~F. Evaluating and clustering retrosynthesis pathways with learned strategy. \emph{Chem. Sci.} \textbf{2021}, \emph{12}, 1469--1478\relax
\mciteBstWouldAddEndPuncttrue
\mciteSetBstMidEndSepPunct{\mcitedefaultmidpunct}
{\mcitedefaultendpunct}{\mcitedefaultseppunct}\relax
\EndOfBibitem
\bibitem[Sacha \latin{et~al.}(2021)Sacha, Blaz, Byrski, Dabrowski-Tumanski, Chrominski, Loska, Wlodarczyk-Pruszynski, and Jastrzebski]{Sacha2021molecule}
Sacha,~M.; Blaz,~M.; Byrski,~P.; Dabrowski-Tumanski,~P.; Chrominski,~M.; Loska,~R.; Wlodarczyk-Pruszynski,~P.; Jastrzebski,~S. Molecule Edit Graph Attention Network: Modeling Chemical Reactions as Sequences of Graph Edits. \emph{J. Chem. Inf. Model.} \textbf{2021}, \emph{61}, 3273--3284, PMID: 34251814\relax
\mciteBstWouldAddEndPuncttrue
\mciteSetBstMidEndSepPunct{\mcitedefaultmidpunct}
{\mcitedefaultendpunct}{\mcitedefaultseppunct}\relax
\EndOfBibitem
\bibitem[Irwin \latin{et~al.}(2022)Irwin, Dimitriadis, He, and Bjerrum]{irwin2022chemformer}
Irwin,~R.; Dimitriadis,~S.; He,~J.; Bjerrum,~E.~J. Chemformer: a pre-trained transformer for computational chemistry. \emph{Mach. Learn. Sci.Technol.} \textbf{2022}, \emph{3}, 015022\relax
\mciteBstWouldAddEndPuncttrue
\mciteSetBstMidEndSepPunct{\mcitedefaultmidpunct}
{\mcitedefaultendpunct}{\mcitedefaultseppunct}\relax
\EndOfBibitem
\bibitem[Maziarz \latin{et~al.}(2024)Maziarz, Liu, Misztela, Kornev, Gai{\'n}ski, Hoefling, Fortunato, Gupta, and Segler]{maziarz2024chimera}
Maziarz,~K.; Liu,~G.; Misztela,~H.; Kornev,~A.; Gai{\'n}ski,~P.; Hoefling,~H.; Fortunato,~M.; Gupta,~R.; Segler,~M. Chimera: Accurate retrosynthesis prediction by ensembling models with diverse inductive biases. \emph{arXiv preprint arXiv:2412.05269} \textbf{2024}, \relax
\mciteBstWouldAddEndPunctfalse
\mciteSetBstMidEndSepPunct{\mcitedefaultmidpunct}
{}{\mcitedefaultseppunct}\relax
\EndOfBibitem
\bibitem[Segler \latin{et~al.}(2018)Segler, Preuss, and Waller]{segler2018planning}
Segler,~M.~H.; Preuss,~M.; Waller,~M.~P. Planning chemical syntheses with deep neural networks and symbolic AI. \emph{Nature} \textbf{2018}, \emph{555}, 604--610\relax
\mciteBstWouldAddEndPuncttrue
\mciteSetBstMidEndSepPunct{\mcitedefaultmidpunct}
{\mcitedefaultendpunct}{\mcitedefaultseppunct}\relax
\EndOfBibitem
\bibitem[Grzybowski \latin{et~al.}(2018)Grzybowski, Szymku{\'c}, Gajewska, Molga, Dittwald, Wo{\l}os, and Klucznik]{grzybowski2018chematica}
Grzybowski,~B.~A.; Szymku{\'c},~S.; Gajewska,~E.~P.; Molga,~K.; Dittwald,~P.; Wo{\l}os,~A.; Klucznik,~T. Chematica: a story of computer code that started to think like a chemist. \emph{Chem} \textbf{2018}, \emph{4}, 390--398\relax
\mciteBstWouldAddEndPuncttrue
\mciteSetBstMidEndSepPunct{\mcitedefaultmidpunct}
{\mcitedefaultendpunct}{\mcitedefaultseppunct}\relax
\EndOfBibitem
\bibitem[Chen \latin{et~al.}(2020)Chen, Li, Dai, and Song]{chen2020retro}
Chen,~B.; Li,~C.; Dai,~H.; Song,~L. Retro*: Learning Retrosynthetic Planning with Neural Guided A* Search. 2020; \url{https://arxiv.org/abs/2006.15820}\relax
\mciteBstWouldAddEndPuncttrue
\mciteSetBstMidEndSepPunct{\mcitedefaultmidpunct}
{\mcitedefaultendpunct}{\mcitedefaultseppunct}\relax
\EndOfBibitem
\bibitem[Genheden \latin{et~al.}(2020)Genheden, Thakkar, Chadimov{\'a}, Reymond, Engkvist, and Bjerrum]{genheden2020aizynthfinder}
Genheden,~S.; Thakkar,~A.; Chadimov{\'a},~V.; Reymond,~J.-L.; Engkvist,~O.; Bjerrum,~E. {AiZynthFinder}: a fast, robust and flexible open-source software for retrosynthetic planning. \emph{J. Cheminf.} \textbf{2020}, \emph{12}, 1--9\relax
\mciteBstWouldAddEndPuncttrue
\mciteSetBstMidEndSepPunct{\mcitedefaultmidpunct}
{\mcitedefaultendpunct}{\mcitedefaultseppunct}\relax
\EndOfBibitem
\bibitem[Molga \latin{et~al.}(2021)Molga, Szymkuć, and Grzybowski]{molga2021chemist}
Molga,~K.; Szymkuć,~S.; Grzybowski,~B.~A. Chemist Ex Machina: Advanced Synthesis Planning by Computers. \emph{Acc. Chem. Res.} \textbf{2021}, \emph{54}, 1094--1106\relax
\mciteBstWouldAddEndPuncttrue
\mciteSetBstMidEndSepPunct{\mcitedefaultmidpunct}
{\mcitedefaultendpunct}{\mcitedefaultseppunct}\relax
\EndOfBibitem
\bibitem[Genheden and Bjerrum(2022)Genheden, and Bjerrum]{genheden_bjerrum_2022}
Genheden,~S.; Bjerrum,~E. PaRoutes: a framework for benchmarking retrosynthesis route predictions. \emph{ChemRxiv} \textbf{2022}, \relax
\mciteBstWouldAddEndPunctfalse
\mciteSetBstMidEndSepPunct{\mcitedefaultmidpunct}
{}{\mcitedefaultseppunct}\relax
\EndOfBibitem
\bibitem[Maziarz \latin{et~al.}(2023)Maziarz, Tripp, Liu, Stanley, Xie, Gai{\'n}ski, Seidl, and Segler]{maziarz2023re}
Maziarz,~K.; Tripp,~A.; Liu,~G.; Stanley,~M.; Xie,~S.; Gai{\'n}ski,~P.; Seidl,~P.; Segler,~M. Re-evaluating Retrosynthesis Algorithms with Syntheseus. \emph{arXiv preprint arXiv:2310.19796} \textbf{2023}, \relax
\mciteBstWouldAddEndPunctfalse
\mciteSetBstMidEndSepPunct{\mcitedefaultmidpunct}
{}{\mcitedefaultseppunct}\relax
\EndOfBibitem
\bibitem[Tu \latin{et~al.}(2025)Tu, Choure, Fong, Roh, Levin, Yu, Joung, Morgan, Li, Sun, \latin{et~al.} others]{tu2025askcos}
Tu,~Z.; Choure,~S.~J.; Fong,~M.~H.; Roh,~J.; Levin,~I.; Yu,~K.; Joung,~J.~F.; Morgan,~N.; Li,~S.-C.; Sun,~X.; others ASKCOS: Open-Source, Data-Driven Synthesis Planning. \emph{Accounts of Chemical Research} \textbf{2025}, \emph{58}, 1764--1775\relax
\mciteBstWouldAddEndPuncttrue
\mciteSetBstMidEndSepPunct{\mcitedefaultmidpunct}
{\mcitedefaultendpunct}{\mcitedefaultseppunct}\relax
\EndOfBibitem
\bibitem[Corey(1967)]{corey1967general}
Corey,~E.~J. General methods for the construction of complex molecules. \emph{Pure and Applied chemistry} \textbf{1967}, \emph{14}, 19--38\relax
\mciteBstWouldAddEndPuncttrue
\mciteSetBstMidEndSepPunct{\mcitedefaultmidpunct}
{\mcitedefaultendpunct}{\mcitedefaultseppunct}\relax
\EndOfBibitem
\bibitem[Badowski \latin{et~al.}(2019)Badowski, Molga, and Grzybowski]{badowski2019selection}
Badowski,~T.; Molga,~K.; Grzybowski,~B.~A. Selection of cost-effective yet chemically diverse pathways from the networks of computer-generated retrosynthetic plans. \emph{Chemical science} \textbf{2019}, \emph{10}, 4640--4651\relax
\mciteBstWouldAddEndPuncttrue
\mciteSetBstMidEndSepPunct{\mcitedefaultmidpunct}
{\mcitedefaultendpunct}{\mcitedefaultseppunct}\relax
\EndOfBibitem
\bibitem[Toniato \latin{et~al.}(2023)Toniato, Vaucher, Schwaller, and Laino]{toniato2023enhancing}
Toniato,~A.; Vaucher,~A.~C.; Schwaller,~P.; Laino,~T. Enhancing diversity in language based models for single-step retrosynthesis. \emph{Digital Discovery} \textbf{2023}, \emph{2}, 489--501\relax
\mciteBstWouldAddEndPuncttrue
\mciteSetBstMidEndSepPunct{\mcitedefaultmidpunct}
{\mcitedefaultendpunct}{\mcitedefaultseppunct}\relax
\EndOfBibitem
\bibitem[Thakkar \latin{et~al.}(2023)Thakkar, Vaucher, Byekwaso, Schwaller, Toniato, and Laino]{thakkar2023unbiasing}
Thakkar,~A.; Vaucher,~A.~C.; Byekwaso,~A.; Schwaller,~P.; Toniato,~A.; Laino,~T. Unbiasing retrosynthesis language models with disconnection prompts. \emph{ACS Central Science} \textbf{2023}, \emph{9}, 1488--1498\relax
\mciteBstWouldAddEndPuncttrue
\mciteSetBstMidEndSepPunct{\mcitedefaultmidpunct}
{\mcitedefaultendpunct}{\mcitedefaultseppunct}\relax
\EndOfBibitem
\bibitem[Westerlund \latin{et~al.}(2024)Westerlund, Saigiridharan, and Genheden]{westerlund2024constrained}
Westerlund,~A.~M.; Saigiridharan,~L.; Genheden,~S. Constrained synthesis planning with disconnection-aware transformer and multi-objective search. 2024; \url{https://chemrxiv.org/engage/chemrxiv/article-details/664ee4c291aefa6ce1c4fc8d}\relax
\mciteBstWouldAddEndPuncttrue
\mciteSetBstMidEndSepPunct{\mcitedefaultmidpunct}
{\mcitedefaultendpunct}{\mcitedefaultseppunct}\relax
\EndOfBibitem
\bibitem[Yu \latin{et~al.}(2024)Yu, Roh, Li, Gao, Wang, and Coley]{yu2024double}
Yu,~K.; Roh,~J.; Li,~Z.; Gao,~W.; Wang,~R.; Coley,~C.~W. Double-Ended Synthesis Planning with Goal-Constrained Bidirectional Search. \emph{arXiv preprint arXiv:2407.06334} \textbf{2024}, \relax
\mciteBstWouldAddEndPunctfalse
\mciteSetBstMidEndSepPunct{\mcitedefaultmidpunct}
{}{\mcitedefaultseppunct}\relax
\EndOfBibitem
\bibitem[Armstrong \latin{et~al.}(2024)Armstrong, Joncev, Guo, and Schwaller]{armstrong2024tango}
Armstrong,~D.; Joncev,~Z.; Guo,~J.; Schwaller,~P. Tango*: Constrained synthesis planning using chemically informed value functions. \emph{arXiv preprint arXiv:2412.03424} \textbf{2024}, \relax
\mciteBstWouldAddEndPunctfalse
\mciteSetBstMidEndSepPunct{\mcitedefaultmidpunct}
{}{\mcitedefaultseppunct}\relax
\EndOfBibitem
\bibitem[Li \latin{et~al.}(2024)Li, Fang, and Lou]{li2024retro}
Li,~J.; Fang,~L.; Lou,~J.-G. Retro-BLEU: quantifying chemical plausibility of retrosynthesis routes through reaction template sequence analysis. \emph{Digital Discovery} \textbf{2024}, \emph{3}, 482--490\relax
\mciteBstWouldAddEndPuncttrue
\mciteSetBstMidEndSepPunct{\mcitedefaultmidpunct}
{\mcitedefaultendpunct}{\mcitedefaultseppunct}\relax
\EndOfBibitem
\bibitem[Kreutter and Reymond(2023)Kreutter, and Reymond]{kreutter2023tripletransformerloop}
Kreutter,~D.; Reymond,~J.-L. Multistep Retrosynthesis combining a disconnection aware triple transformer loop with a route penalty score guided search. \emph{Chemical Science} \textbf{2023}, \emph{14}, 9959--9969\relax
\mciteBstWouldAddEndPuncttrue
\mciteSetBstMidEndSepPunct{\mcitedefaultmidpunct}
{\mcitedefaultendpunct}{\mcitedefaultseppunct}\relax
\EndOfBibitem
\bibitem[Shee \latin{et~al.}(2024)Shee, Li, Morgunov, and Batista]{shee2024directmultistep}
Shee,~Y.; Li,~H.; Morgunov,~A.; Batista,~V. DirectMultiStep: Direct Route Generation for Multi-Step Retrosynthesis. \emph{arXiv preprint arXiv:2405.13983} \textbf{2024}, \relax
\mciteBstWouldAddEndPunctfalse
\mciteSetBstMidEndSepPunct{\mcitedefaultmidpunct}
{}{\mcitedefaultseppunct}\relax
\EndOfBibitem
\bibitem[Xuan-Vu \latin{et~al.}(2025)Xuan-Vu, Armstrong, Joncev, and Schwaller]{xuan2025tempre}
Xuan-Vu,~N.; Armstrong,~D.; Joncev,~Z.; Schwaller,~P. TempRe: Template generation for single and direct multi-step retrosynthesis. \emph{arXiv preprint arXiv:2507.21762} \textbf{2025}, \relax
\mciteBstWouldAddEndPunctfalse
\mciteSetBstMidEndSepPunct{\mcitedefaultmidpunct}
{}{\mcitedefaultseppunct}\relax
\EndOfBibitem
\bibitem[Roh \latin{et~al.}(2025)Roh, Joung, Yu, Tu, Bartholomew, Santiago-Reyes, Fong, Sarpong, Reisman, and Coley]{roh2025higher}
Roh,~J.; Joung,~J.~F.; Yu,~K.; Tu,~Z.; Bartholomew,~G.~L.; Santiago-Reyes,~O.~A.; Fong,~M.~H.; Sarpong,~R.; Reisman,~S.~E.; Coley,~C.~W. Higher-level strategies for computer-aided retrosynthesis. \textbf{2025}, \relax
\mciteBstWouldAddEndPunctfalse
\mciteSetBstMidEndSepPunct{\mcitedefaultmidpunct}
{}{\mcitedefaultseppunct}\relax
\EndOfBibitem
\bibitem[Jablonka \latin{et~al.}(2024)Jablonka, Schwaller, {Ortega-Guerrero}, and Smit]{jablonka2023gpt}
Jablonka,~K.~M.; Schwaller,~P.; {Ortega-Guerrero},~A.; Smit,~B. Leveraging Large Language Models for Predictive Chemistry. \emph{Nat. Mach. Intell.} \textbf{2024}, \emph{6}, 161--169\relax
\mciteBstWouldAddEndPuncttrue
\mciteSetBstMidEndSepPunct{\mcitedefaultmidpunct}
{\mcitedefaultendpunct}{\mcitedefaultseppunct}\relax
\EndOfBibitem
\bibitem[Guo \latin{et~al.}(2023)Guo, Nan, Liang, Guo, Chawla, Wiest, Zhang, \latin{et~al.} others]{llms_for_chem_Guo}
Guo,~T.; Nan,~B.; Liang,~Z.; Guo,~Z.; Chawla,~N.; Wiest,~O.; Zhang,~X.; others What can large language models do in chemistry? a comprehensive benchmark on eight tasks. \emph{Advances in Neural Information Processing Systems} \textbf{2023}, \emph{36}, 59662--59688\relax
\mciteBstWouldAddEndPuncttrue
\mciteSetBstMidEndSepPunct{\mcitedefaultmidpunct}
{\mcitedefaultendpunct}{\mcitedefaultseppunct}\relax
\EndOfBibitem
\bibitem[Bran \latin{et~al.}(2024)Bran, Cox, Schilter, Baldassari, White, and Schwaller]{bran2023chemcrow}
Bran,~A.; Cox,~S.; Schilter,~O.; Baldassari,~C.; White,~A.~D.; Schwaller,~P. Augmenting large language models with chemistry tools. \emph{Nature Machine Intelligence} \textbf{2024}, 1--11\relax
\mciteBstWouldAddEndPuncttrue
\mciteSetBstMidEndSepPunct{\mcitedefaultmidpunct}
{\mcitedefaultendpunct}{\mcitedefaultseppunct}\relax
\EndOfBibitem
\bibitem[Boiko \latin{et~al.}(2023)Boiko, MacKnight, Kline, and Gomes]{boiko2023autonomous}
Boiko,~D.~A.; MacKnight,~R.; Kline,~B.; Gomes,~G. Autonomous chemical research with large language models. \emph{Nature} \textbf{2023}, \emph{624}, 570--578\relax
\mciteBstWouldAddEndPuncttrue
\mciteSetBstMidEndSepPunct{\mcitedefaultmidpunct}
{\mcitedefaultendpunct}{\mcitedefaultseppunct}\relax
\EndOfBibitem
\bibitem[Jablonka \latin{et~al.}(2023)Jablonka, Ai, Al-Feghali, Badhwar, Bocarsly, Bran, Bringuier, Brinson, Choudhary, Circi, \latin{et~al.} others]{jablonka202314}
Jablonka,~K.~M.; Ai,~Q.; Al-Feghali,~A.; Badhwar,~S.; Bocarsly,~J.~D.; Bran,~A.~M.; Bringuier,~S.; Brinson,~L.~C.; Choudhary,~K.; Circi,~D.; others 14 examples of how LLMs can transform materials science and chemistry: a reflection on a large language model hackathon. \emph{Digital discovery} \textbf{2023}, \emph{2}, 1233--1250\relax
\mciteBstWouldAddEndPuncttrue
\mciteSetBstMidEndSepPunct{\mcitedefaultmidpunct}
{\mcitedefaultendpunct}{\mcitedefaultseppunct}\relax
\EndOfBibitem
\bibitem[Rankovi{\'c} and Schwaller(2023)Rankovi{\'c}, and Schwaller]{rankovic2023bochemian}
Rankovi{\'c},~B.; Schwaller,~P. BoChemian: Large Language Model Embeddings for Bayesian Optimization of Chemical Reactions. NeurIPS 2023 Workshop on Adaptive Experimental Design and Active Learning in the Real World. 2023\relax
\mciteBstWouldAddEndPuncttrue
\mciteSetBstMidEndSepPunct{\mcitedefaultmidpunct}
{\mcitedefaultendpunct}{\mcitedefaultseppunct}\relax
\EndOfBibitem
\bibitem[Rankovi{\'c} and Schwaller(2025)Rankovi{\'c}, and Schwaller]{rankovic2025gollum}
Rankovi{\'c},~B.; Schwaller,~P. GOLLuM: Gaussian Process Optimized LLMs--Reframing LLM Finetuning through Bayesian Optimization. \emph{arXiv preprint arXiv:2504.06265} \textbf{2025}, \relax
\mciteBstWouldAddEndPunctfalse
\mciteSetBstMidEndSepPunct{\mcitedefaultmidpunct}
{}{\mcitedefaultseppunct}\relax
\EndOfBibitem
\bibitem[Wang \latin{et~al.}(2025)Wang, Guo, Kong, Ramprasad, Schwaller, Du, and Zhang]{wang2025llm}
Wang,~H.; Guo,~J.; Kong,~L.; Ramprasad,~R.; Schwaller,~P.; Du,~Y.; Zhang,~C. LLM-Augmented Chemical Synthesis and Design Decision Programs. \emph{arXiv preprint arXiv:2505.07027} \textbf{2025}, \relax
\mciteBstWouldAddEndPunctfalse
\mciteSetBstMidEndSepPunct{\mcitedefaultmidpunct}
{}{\mcitedefaultseppunct}\relax
\EndOfBibitem
\bibitem[Song \latin{et~al.}(2025)Song, Pan, Zhao, Ye, Zhang, Tang, and Yu]{song2025aot}
Song,~X.; Pan,~X.; Zhao,~X.; Ye,~H.; Zhang,~S.; Tang,~J.; Yu,~T. AOT*: Efficient Synthesis Planning via LLM-Empowered AND-OR Tree Search. \emph{arXiv preprint arXiv:2509.20988} \textbf{2025}, \relax
\mciteBstWouldAddEndPunctfalse
\mciteSetBstMidEndSepPunct{\mcitedefaultmidpunct}
{}{\mcitedefaultseppunct}\relax
\EndOfBibitem
\bibitem[Schilling-Wilhelmi \latin{et~al.}(2025)Schilling-Wilhelmi, R{\'\i}os-Garc{\'\i}a, Shabih, Gil, Miret, Koch, M{\'a}rquez, and Jablonka]{schilling2025text}
Schilling-Wilhelmi,~M.; R{\'\i}os-Garc{\'\i}a,~M.; Shabih,~S.; Gil,~M.~V.; Miret,~S.; Koch,~C.~T.; M{\'a}rquez,~J.~A.; Jablonka,~K.~M. From text to insight: large language models for chemical data extraction. \emph{Chemical Society Reviews} \textbf{2025}, \relax
\mciteBstWouldAddEndPunctfalse
\mciteSetBstMidEndSepPunct{\mcitedefaultmidpunct}
{}{\mcitedefaultseppunct}\relax
\EndOfBibitem
\bibitem[Narayanan \latin{et~al.}(2025)Narayanan, Braza, Griffiths, Bou, Wellawatte, Ramos, Mitchener, Rodriques, and White]{narayanan2025training}
Narayanan,~S.~M.; Braza,~J.~D.; Griffiths,~R.-R.; Bou,~A.; Wellawatte,~G.; Ramos,~M.~C.; Mitchener,~L.; Rodriques,~S.~G.; White,~A.~D. Training a Scientific Reasoning Model for Chemistry. \emph{arXiv preprint arXiv:2506.17238} \textbf{2025}, \relax
\mciteBstWouldAddEndPunctfalse
\mciteSetBstMidEndSepPunct{\mcitedefaultmidpunct}
{}{\mcitedefaultseppunct}\relax
\EndOfBibitem
\bibitem[Bran \latin{et~al.}(2025)Bran, Neukomm, Armstrong, Jon{\v{c}}ev, and Schwaller]{bran2025chemical}
Bran,~A.~M.; Neukomm,~T.~A.; Armstrong,~D.~P.; Jon{\v{c}}ev,~Z.; Schwaller,~P. Chemical reasoning in LLMs unlocks steerable synthesis planning and reaction mechanism elucidation. \emph{arXiv preprint arXiv:2503.08537} \textbf{2025}, \relax
\mciteBstWouldAddEndPunctfalse
\mciteSetBstMidEndSepPunct{\mcitedefaultmidpunct}
{}{\mcitedefaultseppunct}\relax
\EndOfBibitem
\bibitem[Genheden \latin{et~al.}(2021)Genheden, Engkvist, and Bjerrum]{genheden2021clustering}
Genheden,~S.; Engkvist,~O.; Bjerrum,~E. Clustering of synthetic routes using tree edit distance. \emph{Journal of Chemical Information and Modeling} \textbf{2021}, \emph{61}, 3899--3907\relax
\mciteBstWouldAddEndPuncttrue
\mciteSetBstMidEndSepPunct{\mcitedefaultmidpunct}
{\mcitedefaultendpunct}{\mcitedefaultseppunct}\relax
\EndOfBibitem
\bibitem[Genheden and Shields(2025)Genheden, and Shields]{genheden2025simple}
Genheden,~S.; Shields,~J.~D. A simple similarity metric for comparing synthetic routes. \emph{Digital Discovery} \textbf{2025}, \emph{4}, 46--53\relax
\mciteBstWouldAddEndPuncttrue
\mciteSetBstMidEndSepPunct{\mcitedefaultmidpunct}
{\mcitedefaultendpunct}{\mcitedefaultseppunct}\relax
\EndOfBibitem
\bibitem[Guo \latin{et~al.}()Guo, Kabeshov, Le, Genheden, Bergonzini, Engkvist, and Kaski]{genheden2025evaluation}
Guo,~Y.; Kabeshov,~M.; Le,~T. H.~D.; Genheden,~S.; Bergonzini,~G.; Engkvist,~O.; Kaski,~S. An Expert-Augmented Deep Learning Approach for Synthesis Route Evaluation. \relax
\mciteBstWouldAddEndPunctfalse
\mciteSetBstMidEndSepPunct{\mcitedefaultmidpunct}
{}{\mcitedefaultseppunct}\relax
\EndOfBibitem
\bibitem[Gou \latin{et~al.}(2021)Gou, Yu, Maybank, and Tao]{gou2021knowledge}
Gou,~J.; Yu,~B.; Maybank,~S.~J.; Tao,~D. Knowledge distillation: A survey. \emph{International journal of computer vision} \textbf{2021}, \emph{129}, 1789--1819\relax
\mciteBstWouldAddEndPuncttrue
\mciteSetBstMidEndSepPunct{\mcitedefaultmidpunct}
{\mcitedefaultendpunct}{\mcitedefaultseppunct}\relax
\EndOfBibitem
\bibitem[Chen \latin{et~al.}(2021)Chen, Tworek, Jun, Yuan, Pinto, Kaplan, Edwards, Burda, Joseph, Brockman, \latin{et~al.} others]{chen2021evaluating}
Chen,~M.; Tworek,~J.; Jun,~H.; Yuan,~Q.; Pinto,~H. P. d.~O.; Kaplan,~J.; Edwards,~H.; Burda,~Y.; Joseph,~N.; Brockman,~G.; others Evaluating large language models trained on code. \emph{arXiv preprint arXiv:2107.03374} \textbf{2021}, \relax
\mciteBstWouldAddEndPunctfalse
\mciteSetBstMidEndSepPunct{\mcitedefaultmidpunct}
{}{\mcitedefaultseppunct}\relax
\EndOfBibitem
\bibitem[Li \latin{et~al.}(2023)Li, Allal, Zi, Muennighoff, Kocetkov, Mou, Marone, Akiki, Li, Chim, Liu, Zheltonozhskii, Zhuo, Wang, Dehaene, Davaadorj, Lamy-Poirier, Monteiro, Shliazhko, Gontier, Meade, Zebaze, Yee, Umapathi, Zhu, Lipkin, Oblokulov, Wang, Murthy, Stillerman, Patel, Abulkhanov, Zocca, Dey, Zhang, Fahmy, Bhattacharyya, Yu, Singh, Luccioni, Villegas, Kunakov, Zhdanov, Romero, Lee, Timor, Ding, Schlesinger, Schoelkopf, Ebert, Dao, Mishra, Gu, Robinson, Anderson, Dolan-Gavitt, Contractor, Reddy, Fried, Bahdanau, Jernite, Ferrandis, Hughes, Wolf, Guha, von Werra, and de~Vries]{li2023starcoder}
Li,~R. \latin{et~al.}  StarCoder: may the source be with you! \emph{Preprint at https://arxiv.org/abs/2305.06161} \textbf{2023}, \relax
\mciteBstWouldAddEndPunctfalse
\mciteSetBstMidEndSepPunct{\mcitedefaultmidpunct}
{}{\mcitedefaultseppunct}\relax
\EndOfBibitem
\bibitem[Austin \latin{et~al.}(2021)Austin, Odena, Nye, Bosma, Michalewski, Dohan, Jiang, Cai, Terry, Le, \latin{et~al.} others]{austin2021program}
Austin,~J.; Odena,~A.; Nye,~M.; Bosma,~M.; Michalewski,~H.; Dohan,~D.; Jiang,~E.; Cai,~C.; Terry,~M.; Le,~Q.; others Program synthesis with large language models. \emph{arXiv preprint arXiv:2108.07732} \textbf{2021}, \relax
\mciteBstWouldAddEndPunctfalse
\mciteSetBstMidEndSepPunct{\mcitedefaultmidpunct}
{}{\mcitedefaultseppunct}\relax
\EndOfBibitem
\bibitem[DeepMind(2025)]{google_gemini}
DeepMind,~G. Gemini 2.5: Pushing the Frontier with Advanced Reasoning, Multimodality, Long Context, and Next Generation Agentic Capabilities. \url{https://storage.googleapis.com/deepmind-media/gemini/gemini_v2_5_report.pdf}, 2025; Accessed: 2025-09-23\relax
\mciteBstWouldAddEndPuncttrue
\mciteSetBstMidEndSepPunct{\mcitedefaultmidpunct}
{\mcitedefaultendpunct}{\mcitedefaultseppunct}\relax
\EndOfBibitem
\bibitem[{Anthropic}(2024)]{anthropic2024claude35sonnet}
{Anthropic} Introducing Claude 3.5 Sonnet. \url{https://www.anthropic.com/news/claude-3-5-sonnet}, 2024; Press release announcing Claude 3.5 Sonnet with benchmark results\relax
\mciteBstWouldAddEndPuncttrue
\mciteSetBstMidEndSepPunct{\mcitedefaultmidpunct}
{\mcitedefaultendpunct}{\mcitedefaultseppunct}\relax
\EndOfBibitem
\bibitem[Lowe(2012)]{lowe2012extraction}
Lowe,~D.~M. Extraction of chemical structures and reactions from the literature. Ph.D.\ thesis, University of Cambridge, 2012\relax
\mciteBstWouldAddEndPuncttrue
\mciteSetBstMidEndSepPunct{\mcitedefaultmidpunct}
{\mcitedefaultendpunct}{\mcitedefaultseppunct}\relax
\EndOfBibitem
\bibitem[Schwaller \latin{et~al.}(2019)Schwaller, Laino, Gaudin, Bolgar, Hunter, Bekas, and Lee]{schwaller2019molecular}
Schwaller,~P.; Laino,~T.; Gaudin,~T.; Bolgar,~P.; Hunter,~C.~A.; Bekas,~C.; Lee,~A.~A. Molecular transformer: a model for uncertainty-calibrated chemical reaction prediction. \emph{ACS Cent. Sci.} \textbf{2019}, \emph{5}, 1572--1583\relax
\mciteBstWouldAddEndPuncttrue
\mciteSetBstMidEndSepPunct{\mcitedefaultmidpunct}
{\mcitedefaultendpunct}{\mcitedefaultseppunct}\relax
\EndOfBibitem
\bibitem[OpenAI(2023)]{openai_gpt4}
OpenAI GPT-4 Technical Report. \emph{arXiv preprint arXiv:2303.08774} \textbf{2023}, \relax
\mciteBstWouldAddEndPunctfalse
\mciteSetBstMidEndSepPunct{\mcitedefaultmidpunct}
{}{\mcitedefaultseppunct}\relax
\EndOfBibitem
\bibitem[Dobbelaere \latin{et~al.}(2024)Dobbelaere, Lengyel, Stevens, and Van~Geem]{rxn-insight}
Dobbelaere,~M.~R.; Lengyel,~I.; Stevens,~C.~V.; Van~Geem,~K.~M. Rxn-INSIGHT: fast chemical reaction analysis using bond-electron matrices. \emph{Journal of Cheminformatics} \textbf{2024}, \emph{16}, 37\relax
\mciteBstWouldAddEndPuncttrue
\mciteSetBstMidEndSepPunct{\mcitedefaultmidpunct}
{\mcitedefaultendpunct}{\mcitedefaultseppunct}\relax
\EndOfBibitem
\end{mcitethebibliography}

\appendix

\section{Appendix / supplemental material}

\label{app:route_details}

\begin{table}[H]
\centering
\begin{tabular}{lcc}
\hline
Dataset & Synthetic Routes & Single Step Reactions \\
\hline
Train & 371,613 & 623,594 \\
Validation & 19,526 & 32,420 \\
PaRoutes-N1 & 10,000 & 31,785 \\
PaRoutes-N5 & 10,000 & 38,483 \\
\hline
\end{tabular}
\end{table}

\subsection{Extended Methodology}
\label{app:xtended_methods}
\paragraph{Knowledge Distillation Process}
Our method consists of a multi-step process with five phases: Strategy Extraction and Code Generation, Test and Refactor cycles, and two sequential Judgement and Constrained Refactoring phases.
\newline
\textbf{Strategy Extraction and Code Generation Phase:} The LLM analyses synthetic routes and provides natural language descriptions of the synthetic strategies, before generating corresponding binary Python functions. During this phase, the LLM is provided with comprehensive tools and APIs including the full RDKit documentation, a JSON schema representing the synthesis route data structure, tree traversal functions, and functions for extracting products and reactants from reaction SMILES. Additionally, the model receives lists of functional groups, named reactions and ring structures with corresponding APIs for detecting their presence in molecules or reactions.
\newline
\textbf{Test and Refactor Cycle:} Generated functions are evaluated against ground truth synthesis routes. Functions producing errors or failing to return \texttt{True} enter a three-iteration refactor cycle, where the LLM receives test environment STDOUT, error messages, relevant library documentation and APIs. Invalid functions remaining after three iterations are discarded. This phase yielded 11,926 functions.
\newline
\textbf{Initial Screening:} Valid functions are evaluated on the PaRoutes dataset, with functions flagging more than 40
\newline
\textbf{First Constrained Refactoring and Judgement Phase:} Functions undergo refactoring conducted by Gemini-2.5-Flash. Code modifications are strictly limited to: code removal, condition statement modification, and parameterization of functional group, reaction class and ring structure lists. These constraints are enforced through prompt design with explicit instructions on permitted modifications. The LLM judges functions as 'Bad', 'Good', or 'Perfect' according to specified criteria, with freedom to assess general code quality and chemical meaningfulness. Of the 11,926 functions, 9,522 received quality ratings: 107 Perfect (1.1\%), 526 Good (5.5\%), and 8,806 Bad (92.5\%), with 83 receiving Unknown ratings (0.9\%). Following refactoring, 6,232 functions received improved ratings: 965 Perfect (15.5\%), 4,177 Good (67.0\%), and 1,090 Bad (17.5\%). The 5,142 Good and Perfect functions advanced to the next phase.
\newline
\textbf{Second Constrained Refactoring and Judgement Phase:} The Good and Perfect functions from the previous phase undergo final evaluation by Gemini-2.5-Pro using the same constrained refactoring approach. This yielded 1,093 final functions: 851 Good and 242 Perfect. Details of the prompts and evaluation criteria are provided in the Appendix. Once functions were evaluated on real data, 17 were found to contain Python syntax errors and were discarded, resulting in a final library of 1,076 functions.

\paragraph{Strategy Based Retrieval}
Our strategy retrieval framework is designed to identify synthesis routes that match complex, multi-faceted strategic queries. The user provides a set of natural language queries which describe the route. This summary is then translated by the Query Rewriter LLM into a structured JSON query. This structured query is highly expressive, composed of multiple sub-queries linked by logical operators (AND/OR) and supporting negation. Each sub-query contains both a natural-language description for semantic matching and a set of precise atomic filters (e.g., required named reactions, functional groups, ring systems) that leverage the detailed metadata associated with each function in our library.
Upon receiving a query, the StrategyRetriever first identifies a set of candidate functions for each sub-query through a two-stage filtering process. First, an optional semantic filter narrows the search space by selecting the top N functions whose pre-computed description embeddings have the highest cosine similarity to the sub-query's natural-language embedding. Second, an atomic filter further refines this set by retaining only those functions whose metadata satisfies the query's structured Categorical Checks. Using a pre-computed inverted index, the system then identifies all routes that pass at least one of these candidate functions, forming a pool of potential matches.
Routes in this pool are then scored and ranked. The primary ranking criterion is the Match Count -- the number of sub-queries the route successfully satisfies. For routes with an equal Match Count, a secondary Rank Score is calculated. This score is the average cosine similarity between the embedding of each sub-query's description and the embeddings of the specific functions that passed on the route and fulfilled that sub-query. The final list is sorted first by Match Count and then by Rank Score, ensuring that the most relevant and complete matches are prioritized. Retrieval performance is evaluated using Top-K accuracy, measuring whether the ground-truth route is present within the top K results.

\paragraph{Retrieval detailed results}

Here we include a full set of results from our Top-n ( number of semantic functions retrieved ) and Top-k ( accuracy at the k'th retrieved item ) sweep. In general, we see a large increase in accuracy at all all values of Top-k as the number of semantic matches increases. We note there are some inversions, such as from Top-n = 20 -> Top-n = 25, likely due to incorrectly matched semantic functions coincidentally displaying a strong categorical match and hence being prioritized by the ranking system (Table \ref{tab:accuracy_sweep}).
\begin{table}
\caption{Accuracy (\%) at various Top-N retrieval and Top-K evaluation points.}
\label{tab:accuracy_sweep}
\small
\begin{tabularx}{\columnwidth}{r*{11}{X}}
\toprule
Top-N & Top-1 & Top-2 & Top-3 & Top-4 & Top-5 & Top-10 & Top-15 & Top-20 & Top-30 & Top-40 & Top-50 \\
\midrule
1 & 7.27 & 9.09 & 14.55 & 14.55 & 16.36 & 18.18 & 21.82 & 25.45 & 29.09 & 29.09 & 29.09 \\
3 & 14.55 & 27.27 & 29.09 & 36.36 & 38.18 & 41.82 & 47.27 & 50.91 & 60.00 & 61.82 & 61.82 \\
5 & 27.27 & 29.09 & 34.55 & 36.36 & 36.36 & 52.73 & 60.00 & 61.82 & 65.45 & 65.45 & 65.45 \\
10 & 36.36 & 45.45 & 45.45 & 47.27 & 49.09 & 58.18 & 65.45 & 65.45 & 69.09 & 74.55 & 76.36 \\
15 & 43.64 & 50.91 & 52.73 & 54.55 & 56.36 & 60.00 & 67.27 & 69.09 & 70.91 & 70.91 & 74.55 \\
20 & 43.64 & 52.73 & 54.55 & 54.55 & 58.18 & 61.82 & 61.82 & 63.64 & 69.09 & 74.55 & 76.36 \\
25 & 40.00 & 49.09 & 50.91 & 56.36 & 58.18 & 63.64 & 65.45 & 69.09 & 74.55 & 78.18 & 80.00 \\
30 & 38.18 & 47.27 & 50.91 & 54.55 & 58.18 & 61.82 & 65.45 & 69.09 & 76.36 & 76.36 & 81.82 \\
40 & 40.00 & 45.45 & 50.91 & 54.55 & 60.00 & 61.82 & 67.27 & 74.55 & 83.64 & 83.64 & 83.64 \\
50 & 41.82 & 45.45 & 50.91 & 52.73 & 58.18 & 63.64 & 67.27 & 72.73 & 80.00 & 81.82 & 83.64 \\
75 & 43.64 & 50.91 & 50.91 & 52.73 & 60.00 & 65.45 & 70.91 & 74.55 & 80.00 & 83.64 & 83.64 \\
100 & 50.91 & 56.36 & 56.36 & 63.64 & 63.64 & 70.91 & 70.91 & 76.36 & 80.00 & 81.82 & 83.64 \\
150 & 50.91 & 58.18 & 63.64 & 65.45 & 67.27 & 72.73 & 72.73 & 80.00 & 80.00 & 81.82 & 85.45 \\
200 & 52.73 & 60.00 & 65.45 & 67.27 & 69.09 & 72.73 & 76.36 & 78.18 & 83.64 & 83.64 & 85.45 \\
250 & 50.91 & 58.18 & 63.64 & 67.27 & 67.27 & 70.91 & 76.36 & 76.36 & 81.82 & 83.64 & 85.45 \\
\bottomrule
\end{tabularx}
\end{table}

\subsection{Clustering Details}

\paragraph{Strategy Based Clustering}
\label{app:strategy_clustering}
To understand the emergent strategic archetypes within our generated function library, we performed a clustering analysis on the synthesis routes planned for 5,000 molecules from the ChEMBL dataset, as previously used by \citet{genheden2021clustering}. The process began by applying the full library of 1,076 functions to each successful synthesis route. Prior to evaluation, each route underwent a canonicalization pre-processing step: the synthesis tree was sorted to prioritize the deepest sub-trees, and all reaction SMILES strings were converted from a retrosynthetic to a forward-reaction representation. This initial pass annotated each route with a list of "passing functions"---those which returned True.

Following annotation, we transformed this data into a format suitable for clustering. Each route was converted into a binary feature vector of length 1,076, where each dimension corresponds to a unique function. A value of 1 was assigned if the function passed for a given route, and 0 otherwise. We then employed the K-Means clustering algorithm on these vectors. The optimal number of clusters, $k$, was determined by maximizing the silhouette score over a range of potential $k$ values, ensuring a mathematically robust partitioning of the strategic space. Finally, to characterize the nature of each cluster, we identified its defining functions by calculating a "distinctiveness score" for each function. This score is defined as the frequency of the function passing for routes within the cluster minus its frequency for all routes outside the cluster. Functions with high positive scores are strong indicators of a cluster's identity, representing a coherent set of tactical checks that define a particular synthetic approach.

\begin{figure}[t]
    \centering
    \begin{subfigure}{\columnwidth}
        \centering
        \includegraphics[width=\linewidth]{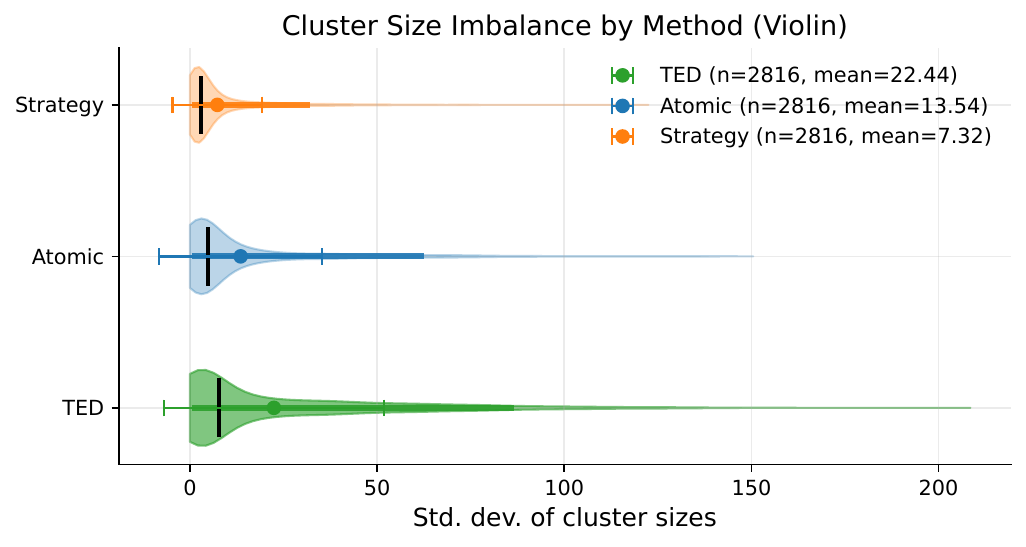}
        \caption{Distribution of cluster size standard deviations.}
        \label{fig:cluster_balance}
    \end{subfigure}
    \vfill 
    \begin{subfigure}{\columnwidth}
        \centering
        \includegraphics[width=\linewidth]{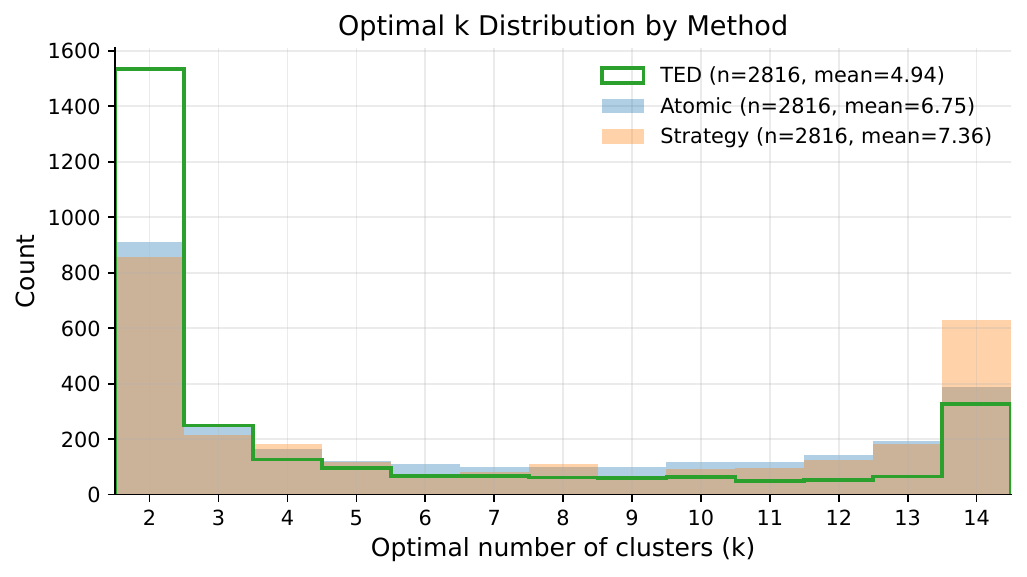}
        \caption{Distribution of the optimal number of clusters ($k$).}
        \label{fig:optimal_k}
    \end{subfigure}
    
    \caption{
        Quantitative comparison of clustering granularity and balance between Strategy-based, Atomic, and Tree Edit Distance (TED) methods.
        \textbf{(a)} The violin plot illustrates the imbalance in cluster sizes. TED (green) exhibits a high mean standard deviation ($\mu=22.44$), indicating it often creates lopsided partitions (e.g., one massive cluster and one outlier). Strategy-based clustering (orange) maintains the lowest deviation ($\mu=7.32$), reflecting balanced grouping.
        \textbf{(b)} The histogram of optimal $k$ values shows that TED strongly prefers coarse clustering (peaking at $k=2$), whereas the Strategy-based method favors higher granularity (mean $k=7.36$), capturing nuanced variations in synthetic approaches.
    }
    \label{fig:clustering_comparison}
\end{figure}

The comparative analysis of clustering outcomes highlights a fundamental difference in how topological versus strategic metrics partition the route space. Figure~\ref{fig:cluster_balance} presents the distribution of standard deviations for cluster sizes across the three methods. Tree Edit Distance (TED) exhibits a notably high mean standard deviation of 22.44. This high variance confirms that TED tends to produce highly imbalanced clusters; specifically, it frequently partitions routes into one dominant cluster containing the vast majority of solutions and a secondary cluster containing a single or very few routes. In contrast, the Strategy-based method demonstrates a significantly lower mean standard deviation of 7.32, with a tighter distribution. This indicates that our feature-based approach identifies groups of comparable size, successfully distinguishing between distinct but equally populated strategic archetypes rather than simply isolating outliers.

Figure~\ref{fig:optimal_k} in the Appendix further elucidates the granularity of these classifications. The distribution of the optimal number of clusters ($k$) for TED is heavily skewed toward low values, with a massive peak at $k=2$ and a mean of 4.94. This suggests that TED lacks the resolution to distinguish fine-grained differences, often defaulting to a binary split of the data. Conversely, the Strategy-based clustering displays a bimodal distribution with a higher mean $k$ of 7.36  and a significant accumulation of cases at the upper bound ($k=14$). This shift toward higher $k$ values demonstrates that the strategic fingerprint provides a richer, more granular vocabulary for classification, allowing for the detection of nuanced differences in synthetic logic that purely topological metrics overlook.

\subsection{Examples of final functions}
\begin{figure}
\centering
   \includegraphics[width=1.0\textwidth]{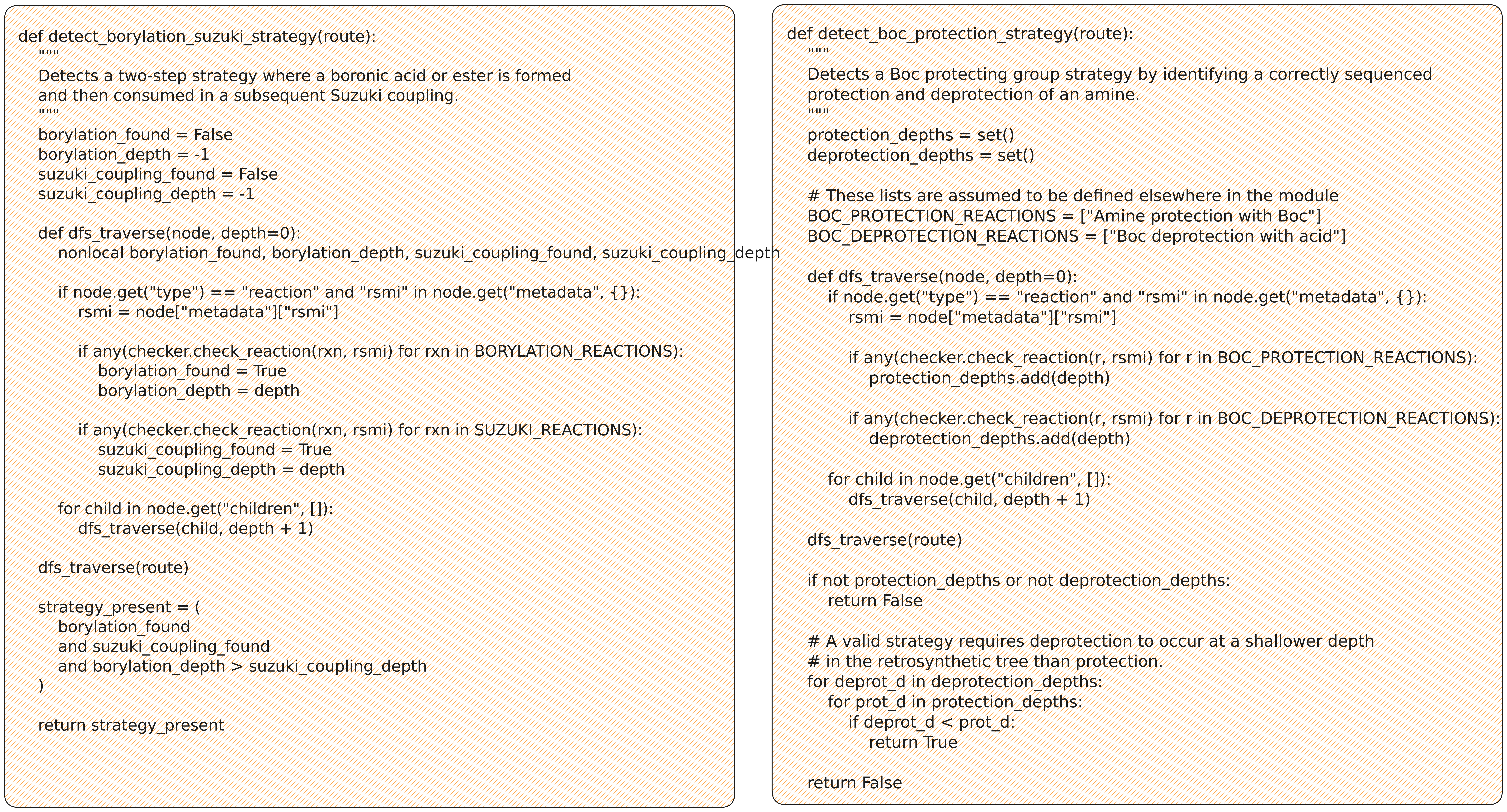}
   \caption{Here we show two further examples of functions evaluated to be "Perfect" by the final LLM Judge. The code has been cleaned by removal of comments, print statements and changing variable names to improve human readability.}
   \label{fig:boc_suzuki}
\end{figure}

\paragraph{Borylation and Suzuki Coupling Strategy}
The Suzuki cross-coupling is a reaction of profound significance in modern organic chemistry, recognized with the 2010 Nobel Prize in Chemistry. Its primary utility lies in the efficient and reliable construction of carbon-carbon bonds, particularly between aromatic rings to form biaryl structures. These biaryl motifs are a cornerstone of countless high-value molecules, including pharmaceuticals (e.g., the anti-inflammatory drug Diflunisal), agrochemicals, and advanced organic materials.

The reaction requires two key partners: an organohalide and an organoboron species (typically a boronic acid or boronic ester). While organohalides are often readily available, the necessary organoboron counterparts for complex targets frequently are not.

This gives rise to a highly powerful and common synthetic plan: a two-step sequence that first creates the organoboron intermediate and then immediately consumes it. The strategy is as follows:
\begin{itemize}
    \item Borylation: A simpler, more accessible precursor molecule is first converted into the required organoboron coupling partner. This step effectively "activates" a C-H or C-Halogen bond by installing the boronic acid/ester "handle."
    \item Suzuki Coupling: This freshly prepared organoboron intermediate is then directly subjected to Suzuki coupling conditions with the second partner, forging the final carbon-carbon bond and completing the construction of the target scaffold.
\end{itemize}

It allows chemists to create carbon-carbon bonds between unsaturated systems in a reaction which is both regio-selective, functional group tolerant and high yielding under standard conditions.

\paragraph{Boc protecting group strategy}
The tert-butoxycarbonyl group, universally known as the Boc group, is one of the most fundamental and widely employed protecting groups in organic synthesis, particularly for amines. Amines are highly prevalent in natural products, pharmaceuticals, and are the building blocks of peptides and proteins. However, their inherent nucleophilicity and basicity mean they will react with a vast array of reagents, often undesirably.

The Boc group provides a robust solution by converting the reactive amine (-NH2) into a stable, unreactive carbamate. This strategic masking is critically important in areas like:

Peptide Synthesis: When coupling amino acids to form a peptide chain, the amine of one amino acid must be protected to prevent it from self-reacting or interfering with the desired amide bond formation. The Boc group was a cornerstone of early solid-phase peptide synthesis.
Multi-step Synthesis: In the construction of complex nitrogen-containing molecules, the Boc group allows chemists to perform reactions on other parts of the molecule (e.g., with strong bases or nucleophiles) without affecting the sensitive amine functionality.
A key advantage of the Boc group is its unique cleavage condition. It is stable to a wide range of reagents but can be removed cleanly and efficiently under mild acidic conditions (e.g., with trifluoroacetic acid, TFA). This selective removal, or "deprotection," allows the amine to be regenerated at the desired stage of a synthesis, making the Boc group an indispensable tool for achieving chemical selectivity.

\section{Constrained Refactoring and Judging Frameworks}
\label{app:refactoring_frameworks}

Here we provide details about the constrained refactoring and LLM judging frameworks used in our evaluation. These frameworks define the rules for both minimal and enhanced code modifications, along with the quality classification systems for judging the outcomes.

\noindent\leaders\hbox to 5pt{\hss-\hss}\hfill\kern0pt
\subsection{Gemini-Flash Refactoring Framework}
The Gemini-Flash evaluation system employs a minimalist approach with strict limitations on permitted modifications.

\paragraph{Allowed Improvements (Strictly Limited)}
The system permits only three types of modifications:
\begin{enumerate}[label=\arabic*.]
    \item \textbf{Code Removal}: Delete buggy, inefficient, or redundant lines (especially redundant functional group checks).
    \item \textbf{Single Conditional Modification}: Fix one clearly inverted or incorrect conditional statement without adding new functions or variables.
    \item \textbf{Description Rewriting}: Update docstrings to accurately reflect what the code actually does after fixes.
\end{enumerate}

\begin{tcolorbox}[colback=red!5!white, colframe=red!75!black, title=Critical Constraint]
No other code editing is allowed---no adding lines, changing variable names, or structural modifications.
\end{tcolorbox}

\paragraph{Quality Classification System}
Functions are rated using three strict categories:

\textbf{PERFECT}
\begin{itemize}
    \item \textbf{Code Quality}: Flawless code with robust \texttt{checker} functions.
    \item \textbf{Strategic Value}: Identifies high-level non-obvious synthetic strategies.
    \item \textbf{Description}: Accurate description.
\end{itemize}

\textbf{GOOD}
\begin{itemize}
    \item \textbf{Code Quality}: Correct code primarily using \texttt{checker} functions.
    \item \textbf{Strategic Value}: Identifies valid but common chemical events.
    \item \textbf{Description}: Accurate description.
\end{itemize}

\textbf{BAD} - A function is classified as BAD if it meets \textbf{ANY} of the following criteria:
\begin{itemize}
    \item Contains bugs or logical flaws.
    \item Fails to use \texttt{checker} functions appropriately.
    \item Identifies trivial/meaningless strategies.
\end{itemize}

\paragraph{Key Evaluation Constraints}
\begin{enumerate}[label=\textbullet]
    \item Must use \texttt{checker} functions when they're the superior option.
    \item Functions checking functional group/ring formation must check both reactants \textbf{AND} products (checking only one side is insufficient).
    \item Reactions are always forward-direction (e.g., ester formation, not cleavage).
    \item Synthesis depth matters: depth 1 = late-stage, depth 6 = early-stage.
    \item Linear/convergent synthesis detection is considered trivial and uninteresting.
÷\end{enumerate}

\noindent\leaders\hbox to 5pt{\hss-\hss}\hfill\kern0pt

\subsection{Gemini Pro Enhanced Refactoring Framework}
The Gemini Pro evaluation system provides expanded capabilities with additional specialized refactoring operations while maintaining strict boundaries.

\paragraph{Evaluation Process \& Allowed Improvements}
The evaluator must meticulously analyze the target function's code, description, and chemical strategy. Five types of improvements are permitted, which can occur independently or concurrently:

\textbf{1. Propagating Context (Special Initial Step):} If the target analysis function (e.g., \texttt{dfs\_traverse(node)}) lacks access to \texttt{reaction}, \texttt{depth}, and \texttt{max\_depth}, the first modification should update its signature and the corresponding call site within the wrapper to pass these parameters correctly. This is a permitted and often necessary refactoring.

\textbf{2. Fixing Code by Removal:} Remove any number of lines that are:
\begin{itemize}
    \item \textit{Buggy or Logically Flawed}: Contains clear errors.
    \item \textit{Inefficient or Redundant}: Performs unnecessary checks (common example: checking for functional groups already handled by more specific checks).
    \item \textit{Source of False Positives}: Uses overly broad or non-specific conditions that incorrectly flag reactions (very common with chemically incorrect and overly permissive FG pattern checks).
\end{itemize}

\textbf{3. Fixing Code by Conditional Modification:} Modify a single conditional statement (\texttt{if condition:}) if its logic is clearly inverted or incorrect. \textit{Constraint}: Do not add new checker functions or variables---only correct existing logic using existing elements.

\textbf{4. Refactoring for Enumeration (Special Case):} This powerful modification is \textbf{ONLY} allowed when:
\begin{itemize}
    \item \textit{Trigger}: The function description specifies a single, specific chemical entity (e.g., ``Checks for pyrrole formation''), but the code implementation checks an explicit, well-defined list of related entities (e.g., \texttt{check\_ring(..., ['pyridine', 'pyrrole', 'piperidine'])}). This does not apply to broad categories like ``any aromatic ring.''
    \item \textit{Action}:
    \begin{enumerate}
        \item \textit{Isolate the List}: Move the chemical entity strings outside the function definition, creating a module-level constant (e.g., \texttt{HETEROCYCLES\_OF\_INTEREST = [...]}).
        \item \textit{Update the Code}: Modify the internal logic to reference the new module-level list.
        \item \textit{Update the Description}: Rewrite the description as a template stating the general purpose and explicitly referencing the list (e.g., ``Checks for the formation of specific heterocyclic rings, including pyridine, pyrrole, and piperidine.'').
    \end{enumerate}
\end{itemize}

\textbf{5. Fixing a Flawed Description:} If the docstring/description is inaccurate or becomes inaccurate after code fixes, \textbf{REWRITE THE DESCRIPTION} to be concise, precise, and perfectly reflect what the final, corrected code actually does.

\begin{tcolorbox}[colback=red!5!white, colframe=red!75!black, title=Critical Constraint]
Any form of code editing not explicitly defined above is \textbf{STRICTLY FORBIDDEN}. Do not add new functions, change variable/function names, or alter fundamental control flow except as required by the rules above.
\end{tcolorbox}

\paragraph{Quality Classification System}
Functions are rated using three strict categories with enhanced precision:

\textbf{PERFECT:}
\begin{itemize}
    \item \textit{Code Quality}: Flawless, robust, efficiently uses \texttt{checker} functions. Logic is chemically and computationally sound, handles edge cases, and cannot generate false positives.
    \item \textit{Strategic Value}: Identifies a high-level, non-obvious synthetic strategy (e.g., chemoselectivity, late-stage functionalization).
    \item \textit{Description}: Accurately reflects the code's function.
\end{itemize}

\textbf{GOOD:}
\begin{itemize}
    \item \textit{Code Quality}: Robust, correct, primarily uses \texttt{checker} functions. Clever and robust non-checkers for structural features might be acceptable. Extremely rare edge cases may be acceptable.
    \item \textit{Strategic Value}: Identifies a valid but common or lower-level chemical event (e.g., a standard named reaction).
    \item \textit{Description}: Accurate.
\end{itemize}

\textbf{BAD:} A function is classified as BAD if it meets \textbf{ANY} of the following:
\begin{itemize}
    \item \textit{Code Quality}: Buggy, contains a critical logical flaw, or fails to use \texttt{checker} functions when they are the superior option.
    \item \textit{Strategic Value}: The strategy is trivial, useless, scientifically meaningless, or purely topological.
\end{itemize}

\subsection{Shared Critical Chemical Caveats}
Both evaluation frameworks adhere to the following domain-specific constraints:
\begin{itemize}
    \item \textbf{Reaction Direction is FORWARD:} All reactions are forward synthetic steps.
    \item \textbf{Synthesis Stages and Depth:} \texttt{depth = 1} is the FINAL step (late-stage); \texttt{depth = max\_depth} is the FIRST step (early-stage).
    \item \textbf{Checker Hierarchy:} Use of the checker API is strongly preferred over hardcoded SMARTS.
    \item \textbf{Formation/Cleavage Checks:} Must confirm presence/absence on both reactant and product sides.
\end{itemize}

\subsection{Framework Comparison}
The key distinction between the two frameworks lies in their permissible modifications:
\begin{itemize}
    \item \textbf{Gemini-Flash}: A minimal intervention approach with only three basic modification types.
    \item \textbf{Gemini Pro}: An enhanced capability with five modification types, including context propagation and enumeration refactoring, providing more sophisticated code restructuring while maintaining strict boundaries.
\end{itemize}

\end{document}